\documentclass[10pt,twocolumn,letterpaper]{article}

\usepackage{iccv}
\usepackage{times}
\usepackage{epsfig}
\usepackage{graphicx}
\usepackage{amsmath}
\usepackage{amssymb}

\usepackage{graphicx}
\usepackage{booktabs}
\usepackage{multirow}
\usepackage{color,xcolor} 
\usepackage[hypcap=false]{caption}
\usepackage{subcaption}
\usepackage{url} 
\usepackage{rotating}

\usepackage[pagebackref=true,breaklinks=true,letterpaper=true,colorlinks,bookmarks=false]{hyperref}

\usepackage[capitalize]{cleveref}
\crefname{section}{Sec.}{Secs.}
\Crefname{section}{Section}{Sections}
\Crefname{table}{Table}{Tables}
\crefname{table}{Tab.}{Tabs.}

\iccvfinalcopy

\ificcvfinal\fi

\begin{document}

\title{Learning Versatile 3D Shape Generation with Improved AR Models}

\author{Simian Luo$^{1,*}$, Xuelin Qian$^{1,*}$, Yanwei Fu$^{1,\dagger}$, Yinda Zhang$^{2,\dagger}$, Ying Tai$^{3}$\\
Zhenyu Zhang$^{3}$, Chengjie Wang$^{3}$, Xiangyang Xue$^{1}$ \\
$^{1}$Fudan University;~$^{2}$Google;~$^{3}$Tencent Youtu Lab
}

\twocolumn[{
\renewcommand\twocolumn[1][]{#1}
\maketitle

\captionsetup[table]{hypcap=false}
\begin{center}
    \vspace{-0.4in}
    \centering
    \includegraphics[width=16.5cm]{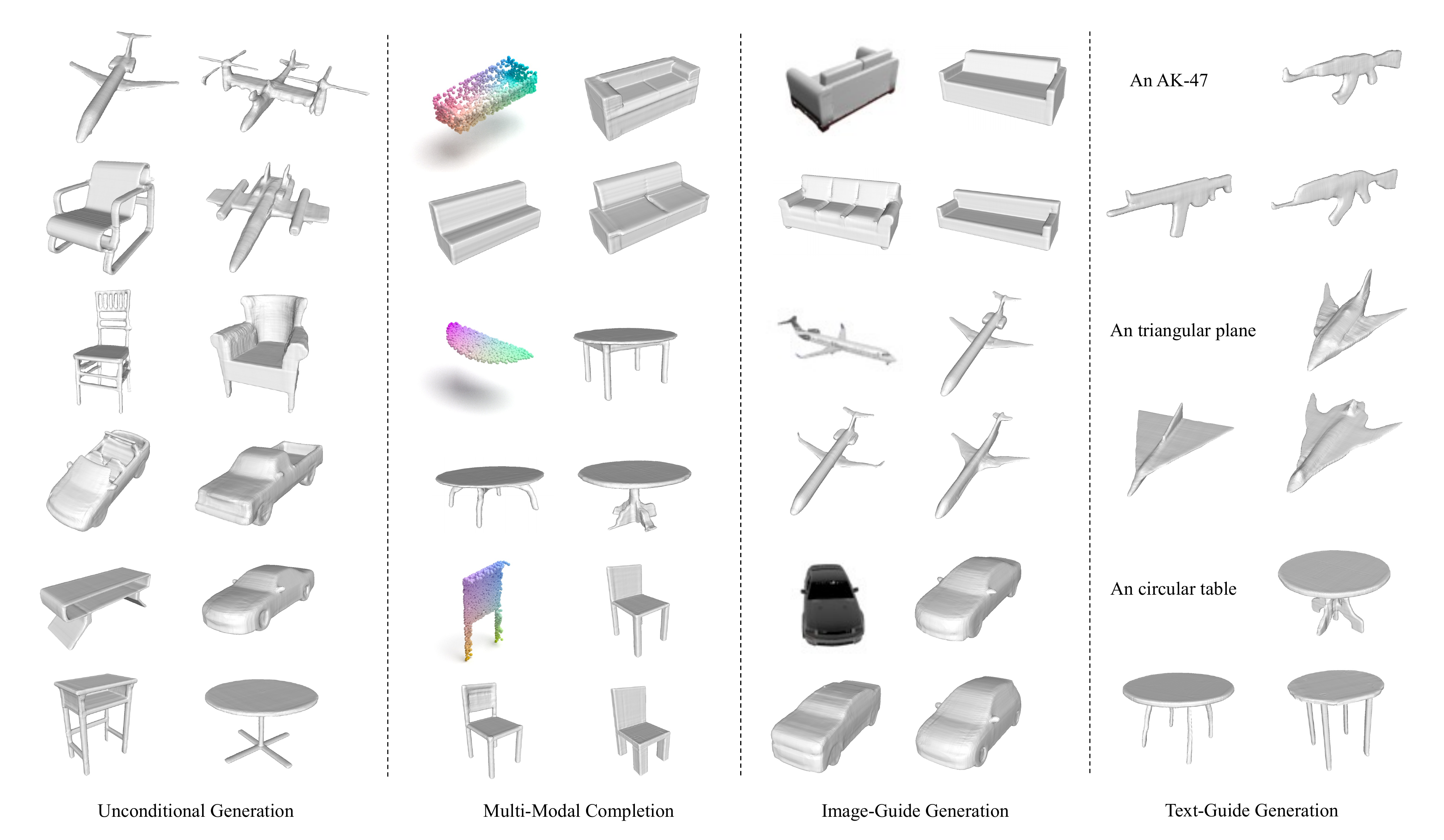}
    \vspace{-0.15in}
    \captionof{figure}{We propose an improved AR model (ImAM) to learn versatile 3D shape generation. ImAM can either generate diverse and faithful shapes with multiple categories via an unconditional way (one column to the left), or can be adapted for conditional generation by incorporating various conditioning inputs given on the left-top (three columns to the right). \label{fig:teaser}} 
    \vspace{-0.05in}
\end{center}}]

\renewcommand{\thefootnote}%
{\fnsymbol{footnote}}
\footnotetext[1]{Equal contribution}
\footnotetext[2]{Corresponding authors}

\begin{abstract}
Auto-Regressive (AR) models have achieved impressive results in 2D image generation by modeling joint distributions in the grid space. While this approach has been extended to the 3D domain for powerful shape generation, it still has two limitations: expensive computations on volumetric grids and ambiguous auto-regressive order along grid dimensions.
To overcome these limitations, we propose the Improved Auto-regressive Model (ImAM) for 3D shape generation, which applies discrete representation learning based on a latent vector instead of volumetric grids. Our approach not only reduces computational costs but also preserves essential geometric details by learning the joint distribution in a more tractable order.
Moreover, thanks to the simplicity of our model architecture, we can naturally extend it from unconditional to conditional generation by concatenating various conditioning inputs, such as point clouds, categories, images, and texts.
Extensive experiments demonstrate that ImAM can synthesize diverse and faithful shapes of multiple categories, achieving state-of-the-art performance.
\end{abstract}

\section{Introduction}

3D shape generation has garnered increasing interest in both academia and industry for its extensive applications in robotics \cite{mees2019self}, autonomous driving \cite{ye2021online,qian2022impdet}, augmented reality \cite{sun2018x} and virtual reality \cite{stets2017visualization}. 
Based on whether user prerequisites are provided, shape generation is typically categorized as unconditional or conditional.
To be an effective generative model, 
it is crucial for the synthesized shapes to be both \textit{diverse} and \textit{faithful}  to the universal cognition of humans or given conditions. These qualities serve as the foundation for other deterministic tasks, such as shape completion, single-view reconstruction, and more.
Previous approaches~\cite{chen2019learning,ibing20213d,park2019deepsdf} usually utilize an AutoEncoder (AE) to learn latent features by shape reconstruction. 
Then, a GAN is trained to fit the distributions of the latent features, allowing for the generation of 3D shapes through  sampling the latent codes learned in AE.
While achieving convincing results, a single embedding for one shape easily encounters the problem of poor scalability and difficulty in training.

Recently, Auto-Regressive (AR) models have shown remarkable performance in the generation of 2D images~\cite{esser2021taming,zhao2021improved,chang2022maskgit} and 3D shape~\cite{mittal2022autosdf,yan2022shapeformer}. Instead of learning a continuous latent space, these model leverage discrete representation learning to encode each 2D/3D input into grid-based discrete codes.
Subsequently, a transformer-based network is employed to jointly model the distribution of all codes, which essentially reflects the underlying prior of objects, facilitating high-quality generation and tractable training.
However, applying AR models to 3D still suffers from two limitations. 
First, as the number of discrete codes increases from squared to cubed, the computational burden of the transformer grows dramatically, making it difficult to converge. 
Second, discrete codes in the grid space are highly coupled. It is ambiguous to simply flatten them for auto-regression (\textit{e.g.}, a top-down row-major order). This may lead to poor quality or even collapse of generated shapes (see Tab.~\ref{tab:ablation_study} and the Supplementary for more details).

In this paper, we propose an improved auto-regressive model (ImAM), to enhance the efficient learning of 3D shape generation. Our key idea is to apply discrete representation learning in a one-dimensional space instead of 3D volumetric space.
Specifically, we first project volumetric grids encoded from 3D shapes onto three axis-aligned orthogonal planes. 
This process significantly  reduces the computational costs from cubed to squared while maintaining the essential information about the input geometry. 
Next, we present a coupling network to further encode three planes into 
a compact and tractable latent space, on which discrete representation learning is performed.

Our ImAM is straightforward and effective, simply tackling the aforementioned limitations by two projections.
Thus, a vanilla decoder-only transformer can be attached to model the joint distributions of codes from the latent spaces. 
Furthermore, the simplicity of the transformer structure allows us to switch freely between unconditional and conditional generation by concatenating various conditioning inputs, such as point clouds, categories, images and texts. 
Figure~\ref{fig:teaser} showcases the ability of our ImAM to generate diverse and accurate shapes across multiple categories, both with and without the given condition on the top-left corner.

In summary, the contributions of this paper are listed as follows.
(1) We propose an improved AR Model (ImAM) for 3D shape generation. By applying discrete representation learning in a latent vector instead of volumetric grids, our ImAM enjoys the advantages of lightweight and flexibility.
(2) Our proposed ImAM model provides 
 a more unified framework for switching between unconditional and conditional generation for a variety of conditioning inputs, including point clouds, categories, images, and texts.
(3) Extensive experiments are conducted on four tasks to demonstrate that our ImAM can generate more faithful and diverse shapes, achieving state-of-the-art results for unconditional and conditional shape generation. 
Overall, our contributions advance the field of 3D shape generation, providing a powerful tool for researchers and practitioners alike.

\section{Related work}
\noindent \textbf{3D shape generative models}.  
As an extremely challenging task, 
we review the most previous efforts by using voxel, point clouds, and implicit representations.
(1) Standard voxel grids can be easily processed by 3D convolution for learning-based 3D task~\cite{maturana2015voxnet,wu20153d,choy20163d,wu2016learning}.
However,  restricted by its cubic space complexity, voxel representation can not scale to a high resolution, usually limited in $64^3$. 
Even with efficient data structures like octrees or multi-scale representation \cite{wang2017cnn,hane2017hierarchical,riegler2017octnetfusion}, 
such representations still have some limitations in quality.
(2)
Point clouds extracted from shape surfaces is an alternative 3D representation \cite{fan2017point,wu2019pointconv,qi2017pointnet}, which  is efficient in terms of memory and does not suffer from the restriction of resolution compared with voxel representations. However, these representation can not represent topological relations in 3D space and are also nontrivial to recover shape surfaces from point cloud representation.
(3)
Recently, implicit  representations have gained attention for simplicity in representing 3D shapes \cite{park2019deepsdf,mescheder2019occupancy,michalkiewicz2019implicit,sitzmann2020metasdf}. By predicting the signed distance \cite{park2019deepsdf,sitzmann2020metasdf} or occupancy label \cite{mescheder2019occupancy,peng2020convolutional} of a given point, and then through Marching cubes \cite{lorensen1987marching} methods, the surface can be easily recovered. Follow-up works \cite{mescheder2019occupancy,park2019deepsdf,chen2019learning,peng2020convolutional,jiang2020local} focus on the design of implicit function representation with global or local shape priors.
To sum up, different 3D representations lead to various shape generative models. There are various good previous works such as 3DGAN \cite{wu2016learning}, PC-GAN \cite{achlioptas2018learning}, IM-GAN \cite{chen2019learning},  and GBIF \cite{ibing20213d}. 
However, most current generative methods are task-specific. And it is difficult to be directly applied to  different generative tasks (\textit{e.g.}, shape completion).
Fundamentally, they rely on GANs for the generation step, suffering from the known drawbacks such as mode collapse and training instability. In contrast, we propose an improved AR model for 3D shape generation that can synthesize high-quality and diverse shapes while being easily generalized to other multi-modal conditions.

\noindent \textbf{Autoregressive models}.
AR models are probabilistic generative approaches that have tractable probability density. Using the probability chain rule, the likelihood of a given sample (usually high dimensional data) can be factorized into a series of product of conditional probability. In contrast, GANs do not have such a tractable probability density.
Recently,  AR models achieve 
remarkable progress  in 2D image generation  \cite{esser2021taming,van2016conditional,razavi2019generating}, to a less extent 3D tasks~\cite{sun2020pointgrow, cheng2022autoregressive, nash2020polygen}.
Most of 3D AR models are struggling with generating high-quality shapes due to the challenges of representation learning with more points or faces.
Particularly, we notice two recent works~\cite{yan2022shapeformer,mittal2022autosdf} that share similar insights of utilizing AR models for 3D tasks. 
\cite{yan2022shapeformer} introduces a sparse representation to only quantize non-empty grids in 3D space, but still follows a monotonic row-major order. \cite{mittal2022autosdf} presents a non-sequential design to break the orders, but performs on all volumetric grids.
However, both of them address only one of the above limitations, and burden the structure design and training of transformer. 
In contrast, our ImAM applies discrete representation learning in a latent vector instead of volumetric grids. Such a representation offers plenty of advantages, including shorter length of discrete codes, tractable orders from auto-regression, fast convergence, and also preserving essential 3D information. Moreover, benefiting from the simplicity of transformers, we can freely switch from unconditional generation to conditional generation by concatenating various conditions.

\begin{figure*} 
\begin{centering}
\includegraphics[scale=0.43]{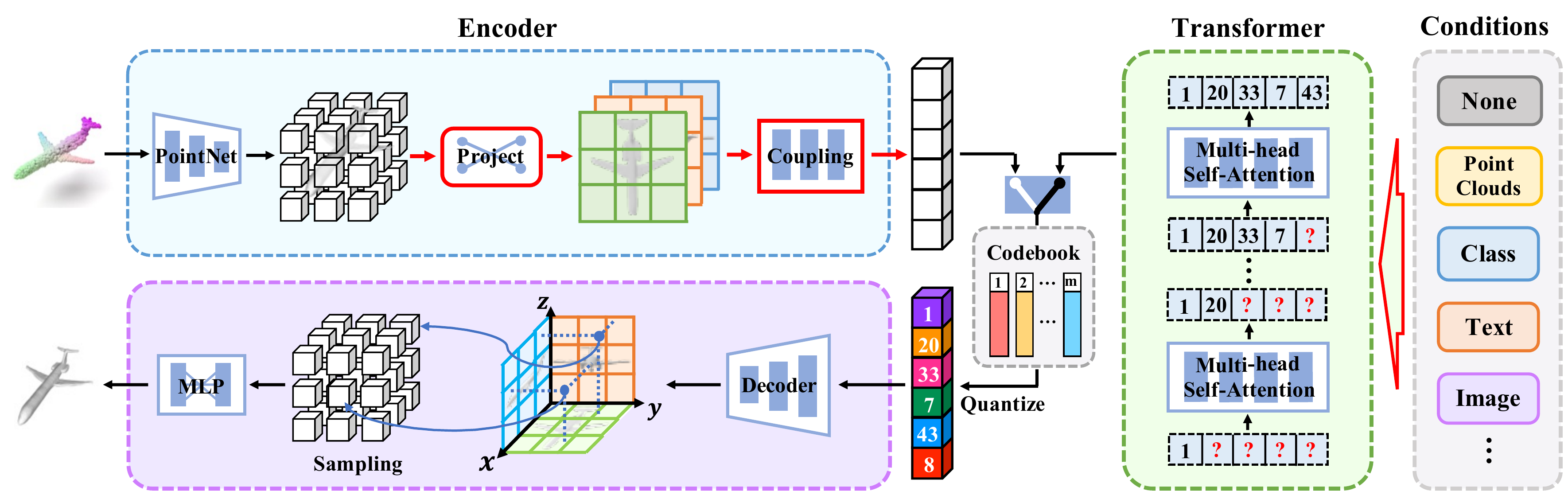}
 \vspace{-0.01in}
\caption{Overview of our ImAM. Given an arbitrary 3D shape, we first project encoded volumetric grids into the three axis-aligned planes, and then use a coupling network to further project them into a latent vector. Vector quantization is thus performed on it for discrete representation. Taking advantages of such a compact representation with tractable orders, vanilla transformers are adopted to auto-repressively learn shape distributions. Furthermore, we can freely switch from unconditional generation to conditional generation by concatenating various conditions, such as point clouds, categories and images. \label{fig:framework}}
 \vspace{-0.2in}
\end{centering}
\end{figure*}

\section{Methodology \label{sec:method}}

Figure~\ref{fig:framework} illustrates the schematic of our proposed framework for 3D shape generation, which consists of a two-stage training procedure. 
It first represents the input as a composition of learned discrete codes (in Sec.~\ref{sec:stage1}), then utilizes a transformer model to learn their interrelations (in Sec.~\ref{sec:stage2}).

\subsection{Improved Discrete Representation Learning \label{sec:stage1}}

\noindent \textbf{Preliminary.}
Given input point clouds $P \in \mathbb{R}^{n\times3}$ where $n$ means the number of points, an encoder is adopted to extract features for each point cloud and then perform voxelization to get features of regular volumetric grids $f^{v} \in \mathbb{R}^{r\times r \times r\times c}$, where $r$ denote the resolution of voxels and $c$ is the feature dimension. To learn discrete representation for each 3D shape, a codebook $\mathbf{q}\in \mathbb{R}^{m\times c}$ is thus introduced whose entry is a learned code describing a particular type of local-part shape in a grid. Formally, for each grid $\left\{ f_{(h,l,w)}^{v} \right\}_{h,l,w=1}^{r}$, vector quantization $\mathcal{Q}\left(\cdot\right)$ is performed by replacing it with the closest entry in codebooks \cite{esser2021taming},
\begin{equation}
   \mathbf{z}^{v} = \mathcal{Q}\left(f^{v}\right) := \arg\min_{\mathbf{e}_{i}\in \mathbf{q}} || f^{v}_{\left(h,l,w\right)} - \mathbf{e}_{i} ||
\label{eq:quantize}
\end{equation}

\noindent where $\mathbf{e}_{i}\in \mathbf{q}$ represents the $i$-th entry in the codebook. 
Thus, learning the correlations between entries in the second stage can explore the underlying priors for shape generation. However, autoregressive generation \cite{esser2021taming,yan2022shapeformer} requires sequential outputs, facing two limitations. \textit{First, the resolution of $\mathbf{z}^{v}$ matters the quality of synthesized shapes}. If $r$ is too small, it lacks the capacity to represent intricate and detailed geometries, while a large value of $r$ can learn a specific code for each local gird, it inevitably increase the computational complexity since the number of required codes explodes as $r$ grows. \textit{Second, the order of $\mathbf{z}^{v}$ affects the generation quality}. Each grid is highly coupled with neighbors, simply flattening (say, along x-y-z axes) can weaken their correlation, leading to sub-optimal generation quality.

One possible solution to solve the first limitation
is applying vector quantization in spatial grids instead of volumetric grids inspired by \cite{chan2021efficient}. Specifically, after obtaining point cloud features $f^{p} \in \mathbb{R}^{n \times c}$, we first project points onto three axis-aligned orthogonal planes. Features of points falling into the same plane grid are aggregated via summation, resulting in three feature maps for the three planes $\left\{ f^{xy}, f^{yz}, f^{xz} \right\} \in \mathbb{R}^{l\times w\times c}$. 
Next, the vector quantization is applied to the three planes separately.
The primary advantage of tri-planar representation is efficient and compact. It can dramatically reduce the number of grids from $\mathcal{O}(r^3)$ to $\mathcal{O}(r^2)$
while preserving essential 3D information. However, it still suffers from the 
the ambiguity of order, 
or worse, since it involves the flattening order of three planes and the order of entries of each plane.

To this end, we further introduce a projection by learning a higher latent space for features of the three planes. This is simply achieved by first concatenating three planes with arbitrary order and then feeding them into a coupling network. Finally, the output is flattened as a projected latent vector, formulated as,

\begin{equation}
f = \tau\left(\mathcal{G}\left(\left[f^{xy};~f^{yz};~f^{xz} \right];~\theta\right)\right) \in \mathbb{R}^{m \times d}
\label{eq:laten_vector}
\end{equation}

\noindent where $\left[\cdot;\cdot\right]$ denotes the concatenation operation; $\mathcal{G}\left(\cdot;\theta\right)$ is a series of convolution layers with parameters $\theta$; $\tau\left(\cdot\right)$ means the operation of flatten with row-major order; $m$ and $d$ indicate the length of latent vector and feature dimension. By applying discrete representation learning in the latent vector, we can describe each 3D shape with $\mathbf{z} = \mathcal{Q}\left(f\right)$, where $\mathcal{Q}\left(\cdot\right)$ represents vector quantization in Eq.~\ref{eq:quantize}.

\noindent \textbf{Remark.} Different from existing works \cite{mittal2022autosdf,yan2022shapeformer} that rely on structure design and training strategies in second stage to address the problem of ambiguous order, we tackle it by learning the coupling relationship of spatial grids in the first stage with the help of the second projection. By stacking convolution layers, we increase the receptive field of each element in the latent vector. Additionally, since the features of each spatial grid on the three planes are fused and highly encoded, each element does not have a definite position mapping in 3D space, which results in a tractable order for auto-regression. More in-depth discussions can be found in Sec.~\ref{sec:more_results}  and the supplementary materials.

\noindent \textbf{Training Objective.} We optimize parameters with reconstruction loss. After getting discrete representation $\mathbf{z}$ which represents indices of entries in the codebook, we retrieve the corresponding codes with indices, denoted as $\mathbf{q}_{\left(\mathbf{z}\right)}$. Subsequently, a decoder with symmetric structures of the encoder are designed to decode $\mathbf{q}_{\left(\mathbf{z}\right)}$ back to features of the three planes \footnote{For more details about the first stage architecture, please refer to the Supplementary.
}. Given sampling points $x\in \mathbb{R}^{3}$, we query their features by projecting them onto each of the three feature planes and performing bilinear interpolation. Features from three planes are accumulated and fed into an implicit function to predict their occupancy values. Finally, we apply binary cross-entropy loss between the predicted values $y_{o}$ and the ground-truth ones $\tilde{y_{o}}$,
\begin{equation}
   \mathcal{L}_{occ} = -\left( \tilde{y_{o}}\cdot\log\left(y_{o}\right) + \left(1-\tilde{y_{o}}\right)\cdot \log\left(1-y_{o}\right)   \right)
\end{equation}

To further train the codebook, we encourage pulling the distance between features before and after the vector quantization. Thus, the codebook loss is derived as,
\begin{equation}
   \mathcal{L}_{code} = \beta||\mathrm{sg}\left[ f\right] - \mathbf{q}_{\left(\mathbf{z}\right)} ||^{2}_{2} +  || f- \mathrm{sg}\left[ \mathbf{q}_{\left(\mathbf{z}\right)}\right]  ||^{2}_{2} 
\end{equation}

\noindent where $\mathrm{sg}\left[\cdot\right]$ denotes the stop-gradient operation \cite{van2017neural} and we set $\beta=0.4$ by default. In sum, the overall loss for the first stage is $\mathcal{L}_{rec}=\mathcal{L}_{occ} + \mathcal{L}_{code}$.

\subsection{Learning Priors with Vanilla Transformers \label{sec:stage2}}

Benefiting from a compact composition and tractable order of discrete representation, models in the second stage can absorbedly learn the correlation between discrete codes,  effectively exploring priors of shape composition. We thus adopt a vanilla decoder-only transformer \cite{esser2021taming} without any specific-designed module.

\textit{For unconditional generation}, given discretized indices of latent vector $\mathbf{z} = \left\{\mathbf{z}_{1}, \mathbf{z}_{2}, \cdots, \mathbf{z}_{m} \right\}$, we feed them into a learnable embedding layer to retrieve features with discrete indices \footnote{We reuse the symbols of $\mathbf{z}$ after embedding for simplicity}. 
Then, the transformer with multi-head self-attention mechanism predicts the next possible index by learning the distribution of previous indices, $p\left(\mathbf{z}_{i}~|~\mathbf{z}_{<i}\right)$. This gives the joint distribution of full representation  as,
\begin{equation}
p\left(\mathbf{z}\right) = \prod_{i=1}^{m}p\left(\mathbf{z}_{i}~|~\mathbf{z}_{<i} \right)
\label{eq:uncond}
\end{equation}

\textit{For conditional generation}, users often expect to control the generation process by providing additional conditions. Instead of designing complex modules or training strategies, we simply learn joint distribution given conditions $\mathbf{c}$ by prepending it to $\mathbf{z}$. Equation~\ref{eq:uncond} is thus  extended as,
\begin{equation}
p\left(\mathbf{z}\right) = \prod_{i=1}^{m}p\left(\mathbf{z}_{i}~|~\mathbf{c},~\mathbf{z}_{<i} \right)
\label{eq:cond}
\end{equation}

\noindent where $\mathbf{c}$ denotes a feature vector of given conditions. The simplicity of our model gives the flexibility to learn conditions of any form. Specifically, for 3D conditions such as point clouds, we use our discrete representation learning in Sec.~\ref{sec:stage1} to transform them into a vector. As for 2D/1D conditions such as images and classes, we either adopt pre-trained models or embedding layers to extract their features.

\noindent \textbf{Objective.}
To train  second stage, we minimize  negative log-likelihood of Eq.~\ref{eq:uncond} or \ref{eq:cond} as $\mathcal{L}_{nll} = \mathbb{E}_{x \sim p\left(x\right)}\left[-\log p\left(\mathbf{z}\right)\right]$, where $p\left(x\right)$ is the distribution of real data.

\noindent \textbf{Inference.} With both models trained on two stages, we use Eq.~\ref{eq:uncond} or \ref{eq:cond} to perform shape generation by progressively sampling the next index with top-\textit{k} sampling strategy, until all elements in $\mathbf{z}$ are completed. Then, we feed $\mathbf{q}_{\left(\mathbf{z}\right)}$ into the decoder of the first stage, and query probabilities of occupancy values for all sampled 3D positions (\textit{e.g., $128^3$}). The output shapes are extracted with Marching Cubes \cite{lorensen1987marching}.

\section{Experiments \label{sec:experiments}}

This section starts with comparing results on unconditional 3D shape generation, showing more faithful and diverse shapes synthesized by our approach in Sec.~\ref{sec:shape_generation}. Next, we show extensive studies on four generation tasks, demonstrating the powerful and flexible ability of ImAM (in Sec.~\ref{sec:class_generation} $\sim$ \ref{sec:text_generation}). Lastly, we provide in-depth studies to evaluate the efficacy of our modules and show generalization to real-world data and zero-shot generation (in Sec.~\ref{sec:more_results}). For all experiments, if necessary, we sample point clouds from output meshes with Poisson Disk Sampling, or reconstruct meshes with our auto-encoder from output points, which is better than Poisson Surface Reconstruction \cite{kazhdan2006poisson}. Please refer to the Supplementary for more details about implementations and qualitative results.

\subsection{Unconditional Shape Generation \label{sec:shape_generation}}

\noindent \textbf{Data.} We consider ShapeNet \cite{chang2015shapenet} as our main dataset for generation, following previous literature \cite{ibing20213d,chen2019learning,sun2020pointgrow}. We use the same training split and evaluation setup from \cite{ibing20213d} for fair comparability. Five categories of car, chair, plane, rifle and table are used for testing. As ground-truth, we extract mesh from voxelized models with $256^3$ resolution in \cite{hane2017hierarchical}.

\noindent \textbf{Baselines.} 
We compare ImAM with five state-of-the-art models, including GAN-based IM-GAN\cite{chen2019learning} and GBIF \cite{ibing20213d}, flow-based PointFlow \cite{yang2019pointflow}, score-based ShapeGF \cite{cai2020learning} and diffusion-based PVD \cite{zhou20213d}. 
We train these methods on the same data split with the official implementation.

\noindent \textbf{Metrics and Settings.} 
The size of the generated set is 5 times the size of the test set, the same as \cite{chen2019learning,ibing20213d}.
As suggested by \cite{chen2019learning}, we use the Light Field Descriptor (LFD) \cite{chen2003visual} as our primary similarity distance metric between two shapes. 
Coverage (COV) \cite{achlioptas2018learning}, Minimum Matching Distance (MMD) \cite{achlioptas2018learning} and Edge Count Difference (ECD) \cite{ibing20213d} are adopted to evaluate the diversity, fidelity and overall quality of synthesized shapes. We also use 1-Nearest Neighbor Accuracy (1-NNA) \cite{yang2019pointflow} (with Chamfer Distance) to measure the distribution similarity. The number of sampled points is 2048. Besides, it is well known that COV does not penalize outliers.
To rule out the false positive coverage, we introduce CovT, counting as match between a generation and ground truth shape only if LFD between them is smaller than a threshold \textit{t}.
In practice, \textit{t} could vary across different categories based on the scale and complexity of the shape, and we empirically use mean MMD as the threshold and found it effective in identifying correct matches.

\begin{table}
\centering
\setlength{\tabcolsep}{1mm}{
\footnotesize
\begin{tabular}{c|l|c|c|c|c|c|c}
\hline
\multirow{2}{*}{\textsc{Metrics}} & \multicolumn{1}{c|}{\multirow{2}{*}{\textsc{Methods}}} & \multicolumn{5}{c|}{\textsc{Categories}} & \multirow{2}{*}{\textsc{Avg}}  \tabularnewline
\cline{3-7}
 & & Plane & Car & Chair & Rifle & Table &  \tabularnewline 
\hline
\hline
\multirow{6}{*}{ECD $\downarrow$} 
& IM-GAN \cite{chen2019learning} & \underline{923} & 3172 & 658 & 371 & 418 & 1108  \tabularnewline 
& GBIF \cite{ibing20213d} & 945 & \underline{2388} & \underline{354} & \underline{195} & 411 & \underline{858} \tabularnewline 
& PointFlow \cite{yang2019pointflow} & 2395 & 5318 & 426 & 2708 & 3559 & 2881 \tabularnewline
& ShapeGF \cite{cai2020learning} & 1200 & 2547 & 443 & 672 & \underline{114} & 915 \tabularnewline
& PVD \cite{zhou20213d} & 6661 & 7404 & 1265 & 3443 & 745 & 3904 \tabularnewline
& \textit{Ours} & \textbf{236} & \textbf{842} & \textbf{27} & \textbf{65} & \textbf{31} & \textbf{240} \tabularnewline  
\hline

\multirow{6}{*}{1-NNA $\downarrow$} 
& IM-GAN \cite{chen2019learning} & 78.18 & 89.39 & 65.83 & 69.38 & 65.31 & 73.62 \tabularnewline
& GBIF \cite{ibing20213d}  & 80.22 & 87.19 & 63.95 & 66.98 & 60.96 & 71.86 \tabularnewline
& PointFlow \cite{yang2019pointflow} & \underline{73.61} & 74.75 & 70.18 & 64.77 & 74.81 & 71.62 \tabularnewline
& ShapeGF \cite{cai2020learning} & 74.72 & \underline{62.81} & \underline{59.15} & \underline{60.65} & \underline{55.58} & \underline{62.58} \tabularnewline
& PVD \cite{zhou20213d} & 81.09 & \textbf{57.37} & 62.36 & 77.32 & 74.31 & 70.49 \tabularnewline
& \textit{Ours} & \textbf{59.95} & 76.58 & \textbf{57.31} & \textbf{57.28} & \textbf{54.76} & \textbf{61.17} \tabularnewline
\hline

\multirow{6}{*}{COV $\uparrow$} 
& IM-GAN \cite{chen2019learning} & 77.01 & 65.37 & 76.38 & 73.21 & \underline{85.71} & 75.53 \tabularnewline
& GBIF \cite{ibing20213d} & \textbf{80.96} & \textbf{78.85} & \textbf{80.95} & \textbf{77.00} & 85.13 & \textbf{80.57} \tabularnewline
& PointFlow \cite{yang2019pointflow} & 65.64 & 64.97 & 57.49 & 48.52 & 71.95 & 61.71 \tabularnewline
& ShapeGF \cite{cai2020learning} & 76.64 & 71.85 & 79.41 & 70.67 & \textbf{87.54} & 77.22 \tabularnewline
& PVD \cite{zhou20213d} & 58.09 & 58.64 & 68.93 & 56.12 & 76.84 & 63.72 \tabularnewline
& \textit{Ours} & \underline{79.11} & \underline{73.25} & \underline{80.81} & \underline{74.26} & 84.01 & \underline{78.29} \tabularnewline
\hline

\multirow{6}{*}{CovT $\uparrow$} 
& IM-GAN \cite{chen2019learning} & \underline{41.03} & 50.63 & \underline{45.68} & \underline{51.68} & 46.50 & 47.10 \tabularnewline
& GBIF \cite{ibing20213d} & 32.38 & 52.76 & 39.77 & 50.00 & 43.68 & 43.72   \tabularnewline
& PointFlow \cite{yang2019pointflow} & 35.85 & 47.76 & 28.48 & 34.81 & 30.98 & 35.57 \tabularnewline
& ShapeGF \cite{cai2020learning} & 40.17 & \underline{53.63} & 43.69 & 51.05 & \textbf{48.50} & \underline{47.41} \tabularnewline
& PVD \cite{zhou20213d} & 12.11 & 43.36 & 38.82 & 33.33 & 43.68 & 34.26 \tabularnewline
& \textit{Ours} & \textbf{45.12} & \textbf{56.64} & \textbf{49.82} & \textbf{55.27} & \underline{48.03} & \textbf{50.98} \tabularnewline
\hline

\multirow{6}{*}{MMD $\downarrow$} 
& IM-GAN \cite{chen2019learning} & \underline{3418} & \underline{1290} & 2881 & \underline{3691} & 2505 & \underline{2757} \tabularnewline
& GBIF \cite{ibing20213d} & 3754 & 1333 & 3015 & 3865 & 2584 & 2910 \tabularnewline
& PointFlow \cite{yang2019pointflow} & 3675 & 1393 & 3322 & 4038 & 2936 & 3072 \tabularnewline
& ShapeGF \cite{cai2020learning} & 3530 & 1307 & \underline{2880} & 3762 & \underline{2420} & 2780 \tabularnewline
& PVD \cite{zhou20213d} & 4376 & 1432 & 3064 & 4274 & 2623 & 3154 \tabularnewline
& \textit{Ours} & \textbf{3124} & \textbf{1213} & \textbf{2703} & \textbf{3628} & \textbf{2374} & \textbf{2608} \tabularnewline 
\hline
\end{tabular}}
\vspace{-0.08in}
\caption{Results of unconditional generation. Models are trained for each category. The best and second results are highlighted in \textbf{bold} and \underline{underlined}. \label{tab:unconditional}}
\vspace{-0.2in}
\end{table}

\begin{figure}
\begin{centering}
\includegraphics[scale=0.4]{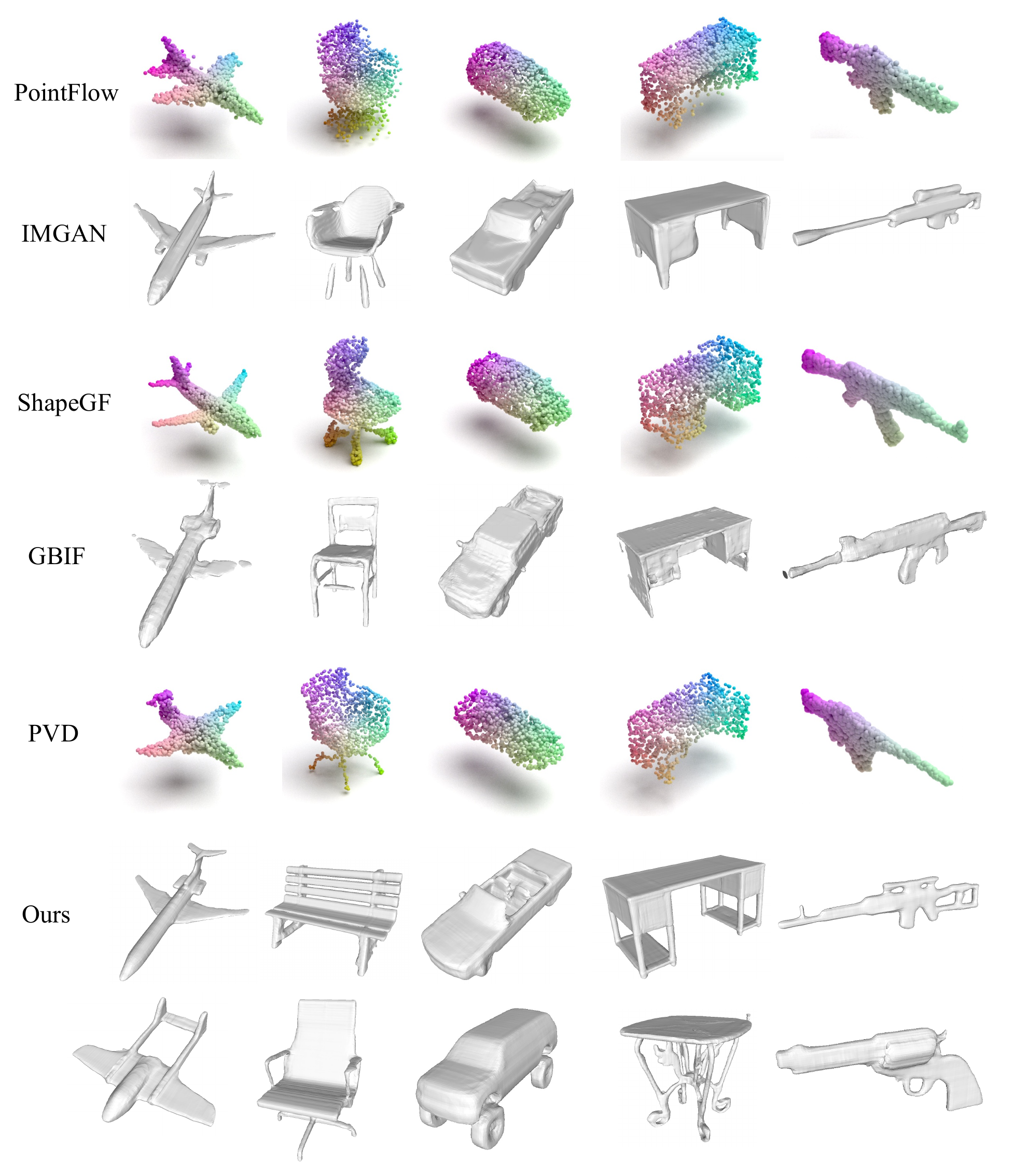}
\vspace{-0.1in}
\caption{Qualitative results of unconditional generation. 
\label{fig:unconditional}}
\vspace{-0.10in}
\end{centering}
\end{figure}

\noindent \textbf{Results Analysis.}
Results are reported in Table~\ref{tab:unconditional}. 
First, ImAM achieves state-of-the-art performance with regard to ECD and 1-NNA. It significantly demonstrates the superiority of our model over synthesizing high-quality shapes. We notice that the result of Car is not good on the metric of 1-NNA. One possible reason is that ImAM tries to generate meshes of tires and seats inside cars, which may not be very friendly to CD.
Second, our model has a clear advantage on both MMD and CovT metrics compared with all competitors, which separately indicates the outstanding fidelity and diversity of our generated shapes.
Third, though GBIF achieves relatively good results on COV, it gets worse results on CovT, suggesting that most of the matched samples come from false positive pairs. 
ImAM, on the contrary, gets second best performance on COV, but higher than GBIF on CovT by about 6 points, 
showing that our generated shapes enjoy the advantage of high-quality and fewer outliers. 
Lastly, we visualize shapes with multiple categories in Fig.~\ref{fig:unconditional}, further supporting the quantitative results and conclusions described above.

\subsection{Class-guide Generation \label{sec:class_generation}}
We first evaluate the versatility of ImAM on class-guide generation, which requires generating shapes given a category label. It is a basic conditional generation task. We use the same dataset and evaluation metrics as in Sec.~\ref{sec:shape_generation}.

\noindent \textbf{Baselines.} We choose two recently-published works as competitors due to the similar motivation. One is two-stage generative model GBIF \cite{ibing20213d}, the other is AR method AutoSDF \cite{mittal2022autosdf}. We simply modify mask-condition of \cite{ibing20213d} to class-condition, and additionally add class token to transformers for \cite{mittal2022autosdf} which is the same as ImAM.

\begin{figure}[t]
\begin{centering}
\includegraphics[width=7.5cm, height=7cm]{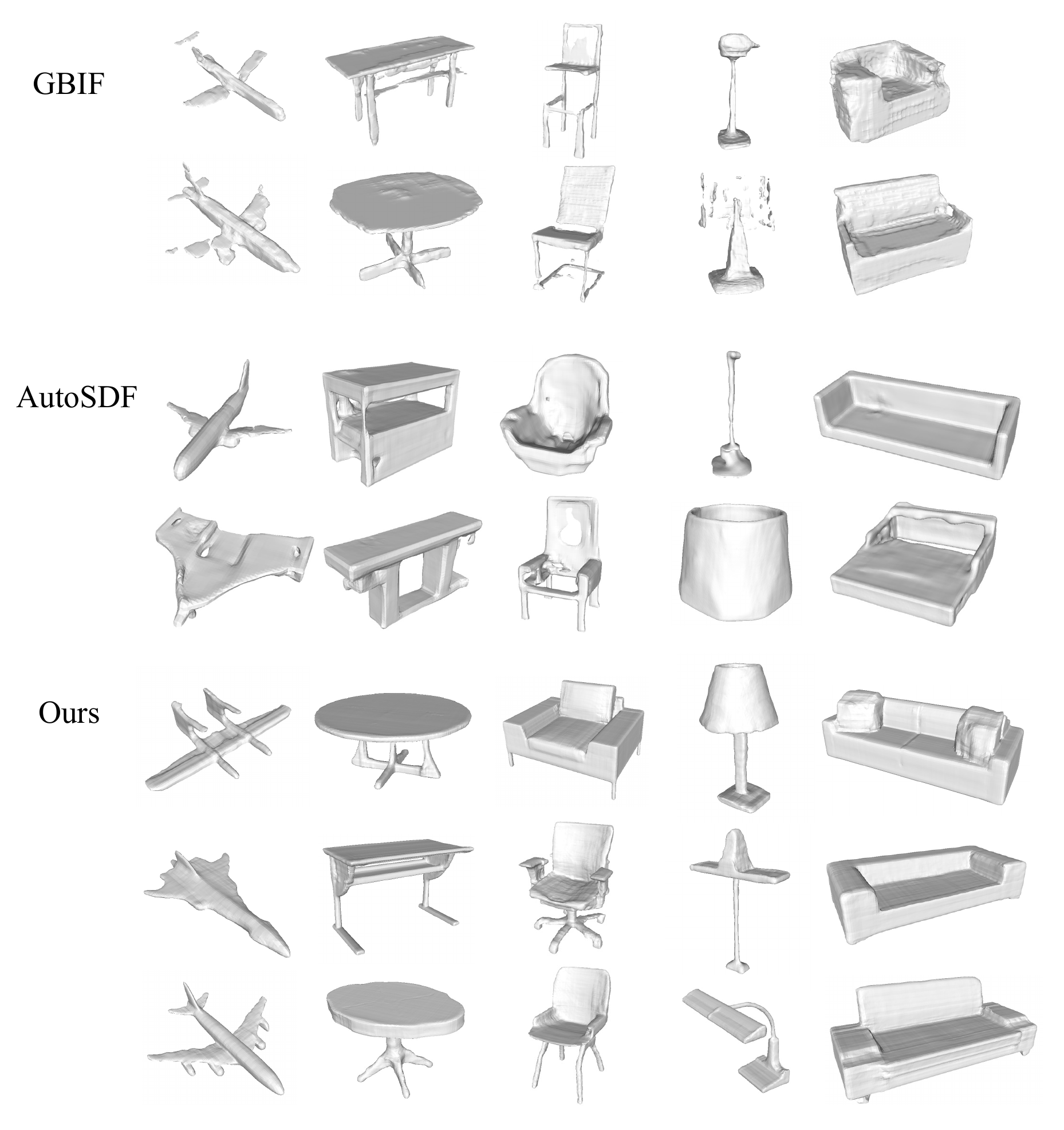}  
\tabularnewline
\vspace{-0.06in}
\caption{Qualitative results of class-guide generation.
\label{fig:class_cond}}
\vspace{-0.08in}
\end{centering}
\end{figure}

\noindent \textbf{Results Analysis.} As shown in Tab.~\ref{tab:class_cond}, ImAM outperforms both competitors across 5 categories by a significant margin. Surprisingly, AutoSDF achieves worse results than GBIF. We explain that the non-sequential auto-regressive way would burden the training of transformer, leading to inferior guidance of simple class tokens. Qualitative results of 5 different categories are further illustrated in Fig.~\ref{fig:class_cond}. As observed, the generated quality of our method is clearly better than GBIF and AutoSDF, while preserving more diversity in types and shapes.

\begin{table}
\centering
\footnotesize 
\setlength{\tabcolsep}{1mm}{
\begin{tabular}{c|l|c|c|c|c|c|c}
\hline
\multirow{2}{*}{\textsc{Metrics}} &  \multicolumn{1}{c|}{\multirow{2}{*}{\textsc{Methods}}} & \multicolumn{5}{c|}{\textsc{Categories}} & \multirow{2}{*}{\textsc{AVG}}  \tabularnewline
\cline{3-7}
 & & Plane & Car & Chair & Rifle  & Table & \tabularnewline 
\hline
\hline
\multirow{3}{*}{COV $\uparrow$} 
& GBIF \cite{ibing20213d} & 68.72 & 69.64 & 75.94 &   68.98 & 81.72 & 73.00 \tabularnewline
& AutoSDF \cite{mittal2022autosdf} & 46.24 & 51.63 & 62.61 &  58.59  & 66.84 & 57.18 \tabularnewline
& \textit{Ours} & \textbf{81.58} & \textbf{71.58} & \textbf{83.98} & \textbf{75.74} & \textbf{85.48} & \textbf{79.67} \tabularnewline

\hline
\multirow{3}{*}{CovT $\uparrow$} 
& GBIF \cite{ibing20213d} & 24.10 & 38.63 & 32.69 & 35.44 & 37.80 & 33.73 \tabularnewline
& AutoSDF \cite{mittal2022autosdf} & 15.43 & 38.03 & 27.82 & 34.60 & 31.22 & 29.42 \tabularnewline
& \textit{Ours} & \textbf{56.49} & \textbf{52.70} & \textbf{45.09} & \textbf{52.74} & \textbf{49.32} & \textbf{51.27} \tabularnewline

\hline
\multirow{3}{*}{MMD $\downarrow$} 
& GBIF \cite{ibing20213d} & 4736 & 1479 & 3220 & 4246 & 2763 & 3289 \tabularnewline
& AutoSDF \cite{mittal2022autosdf} & 5201 & 1477 & 3517 & 4189   & 2992 & 3475 \tabularnewline
& \textit{Ours} & \textbf{3195} & \textbf{1285} & \textbf{2871} & \textbf{3729} & \textbf{2430} & \textbf{2702} \tabularnewline

\hline
\multirow{3}{*}{ECD $\downarrow$} 
& GBIF \cite{ibing20213d} & 1327 & 2752 & 1589 & 434 & 869 & 1394 \tabularnewline
& AutoSDF \cite{mittal2022autosdf} & 5532 & 7352 & 4136 &  2510  &  6354 & 5177 \tabularnewline
& \textit{Ours} & \textbf{571} & \textbf{1889} & \textbf{419} & \textbf{196} & \textbf{285} & \textbf{672} \tabularnewline 
\hline
\end{tabular}}
\vspace{-0.08in}
\caption{Results of class-guide generation. Models are trained on 13 categories of ShapeNet.
\label{tab:class_cond}}
\vspace{-0.17in}
\end{table}

\begin{figure}
 \begin{centering}
\includegraphics[scale=0.3]{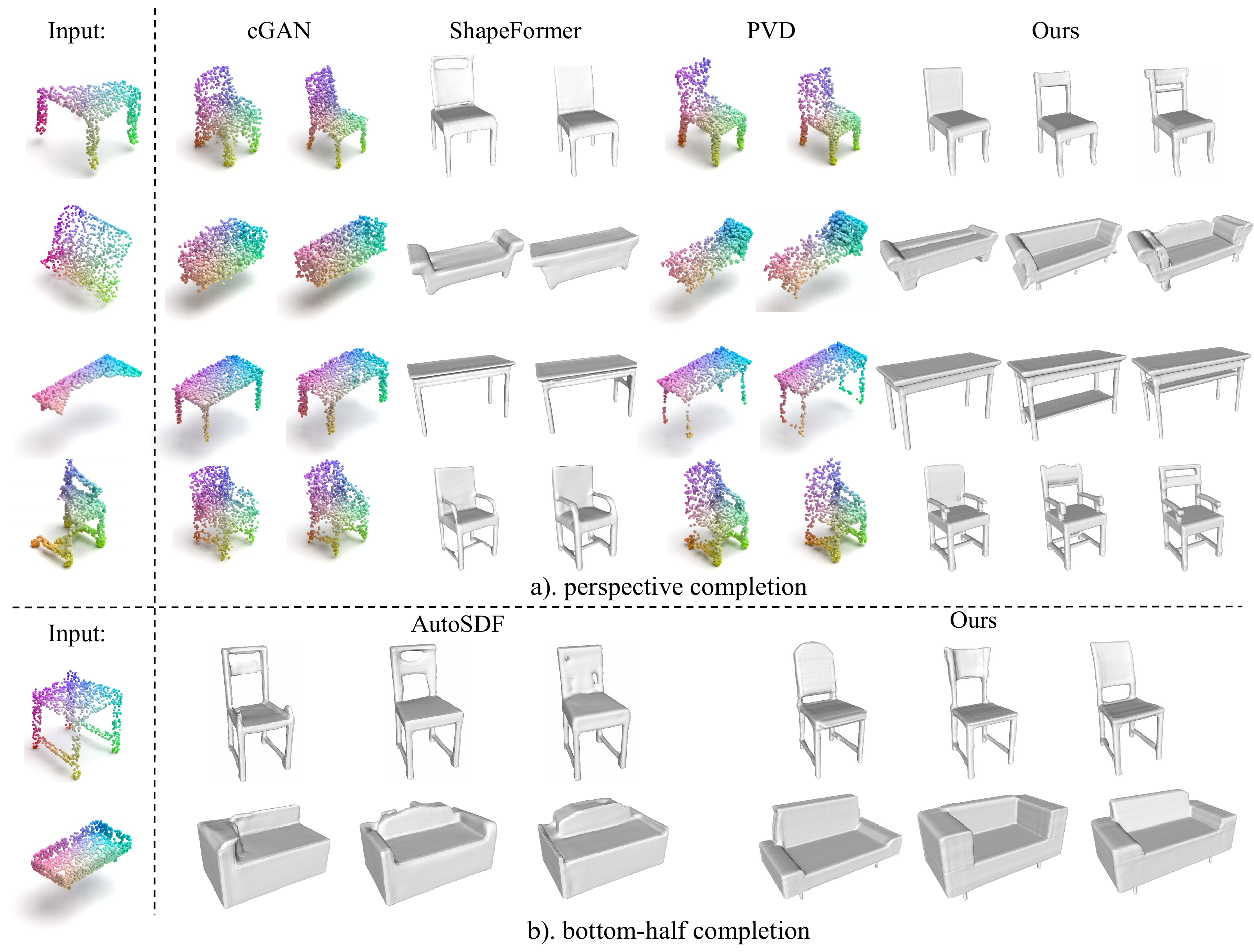}
\vspace{-0.02in}
\caption{Qualitative results of partial point completion. \label{fig:shape_completion}}
\vspace{-0.2in}
\end{centering}
\end{figure}

\subsection{Multi-modal Partial Point Completion}

We further verify the ability of our model in conditional generation by giving partial point clouds. Here, we advocate the multi-modal completion, since there are many possibilities of the completed shape given the partial shape. It is the essence of generative model, where being faithful to the given partial conditions but using your imagination.

\noindent \textbf{Data.} 
We use ShapeNet dataset for testing. Two settings are considered here,
(1) perspective completion \cite{yu2021pointr}: randomly sampling a viewpoint and then removing the $25\% \sim 75\%$ furthest points from the viewpoint;
(2) bottom-half completion \cite{mittal2022autosdf}: removing all points from the top half of shapes.

\noindent \textbf{Baselines.} Four multi-modal completion models are chosen as baselines, including one generative adversarial model cGAN \cite{wu2020multimodal}, one diffusion model PVD \cite{zhou20213d}, two AR models ShapeFormer \cite{yan2022shapeformer} and AutoSDF \cite{mittal2022autosdf}.

\noindent \textbf{Metrics and Settings.} We complete $10$ samples for $100$ randomly selected shapes of three categories, \textit{i.e.}, chair, sofa and table.
Following \cite{wu2020multimodal}, we use Total Mutual Difference (TMD) to measure the diversity. Minimum Matching Distance \cite{achlioptas2018learning} (MMD) with Chamfer Distance and Unidirectional Hausdorff Distance (UHD) \cite{shu20193d} are adopted to measure the faithfulness of completed shapes.

\noindent \textbf{Results Analysis.} We first report perspective completion results in Tab.~\ref{tab:viewpoint_completion}. ImAM beats all baselines and achieves state-of-the-art performance. Importantly, we outperform Shapeformer on all classes and metrics, which also utilizes an AR model with transformers to learn shape distribution. On the other hand, we compare with AutoSDF in its bottom-half completion setting. Results from Tab.~\ref{tab:bottom_completion} illustrates that ImAM outperforms AutoSDF, especially on TMD and UHD. It strongly suggests the flexibility and versatility of our proposed method.
Qualitative results in Fig.~\ref{fig:shape_completion} show the diversity and fidelity of our completed shapes.

\begin{table}
\centering
\footnotesize 
\setlength{\tabcolsep}{2.1mm}{
\begin{tabular}{c|l|c|c|c|c}
\hline
\multirow{2}{*}{\textsc{Metrics}} &  \multicolumn{1}{c|}{\multirow{2}{*}{\textsc{Methods}}} & \multicolumn{3}{c|}{\textsc{Categories}} & \multirow{2}{*}{\textsc{AVG}}  \tabularnewline
\cline{3-5}
 & & Chair & Sofa & Table & \tabularnewline 
\hline
\hline
& cGAN \cite{wu2020multimodal} & 1.708 & 0.687 & \textbf{1.707} & 1.367 \tabularnewline
TMD $\uparrow$ & PVD \cite{zhou20213d} & 1.098 & 0.811 & 0.839 & 0.916  \tabularnewline
($\times 10^{2}$) & ShapeFormer \cite{yan2022shapeformer} & 1.159 & 0.698 & 0.677 & 0.845 \tabularnewline
& \textit{Ours} & \textbf{2.042} & \textbf{1.221} & 1.538 & \textbf{1.600} \tabularnewline 

\hline
& cGAN \cite{wu2020multimodal} & 7.836 & 7.047 & 9.406 & 8.096 \tabularnewline
UHD $\downarrow$ & PVD \cite{zhou20213d} & 10.79 & 13.88 & 11.38 & 12.02 \tabularnewline
($\times 10^{2}$) & ShapeFormer \cite{yan2022shapeformer} & 6.884 & 8.658 & 6.688 & 7.410 \tabularnewline
& \textit{Ours} & \textbf{6.439} & \textbf{6.447} & \textbf{5.948} & \textbf{6.278} \tabularnewline 

\hline
& cGAN \cite{wu2020multimodal} & 1.665 & 1.813 & 1.596 & 1.691 \tabularnewline
MMD $\downarrow$ & PVD \cite{zhou20213d} & 2.352 & 2.041 & 2.174 & 2.189 \tabularnewline
($\times 10^{3}$) & ShapeFormer \cite{yan2022shapeformer} & 1.055 & 1.100 & 1.066 & 1.074 \tabularnewline
& \textit{Ours} & \textbf{0.961} & \textbf{0.819} & \textbf{0.828} & \textbf{0.869} \tabularnewline 
\hline
\end{tabular}}
\vspace{-0.08in}
\caption{Results of multi-modal partial point completion. The missing parts vary according to random viewpoints.}
\label{tab:viewpoint_completion}
\vspace{-0.1in}
\end{table}

\begin{table}
\centering
\footnotesize
\setlength{\tabcolsep}{2.5mm}{
\begin{tabular}{c|l|c|c|c|c} 
\hline
\multirow{2}{*}{\textsc{Metrics}} & \multicolumn{1}{c|}{\multirow{2}{*}{\textsc{Methods}}} & \multicolumn{3}{c|}{\textsc{Categories}} & \multirow{2}{*}{\textsc{AVG}}  \tabularnewline  
\cline{3-5}
&  & Chair & Sofa & Table &  \tabularnewline  
\hline\hline
TMD $\uparrow$ & AutoSDF  \cite{mittal2022autosdf} & 1.230  & 1.422 & 1.381 & 1.344  \tabularnewline 
\multicolumn{1}{c|}{($\times 10^2$)} & \textit{Ours} & \textbf{3.682} & \textbf{2.673} & \textbf{10.30} & \textbf{5.552}  \tabularnewline 
\hline
UHD $\downarrow$ & AutoSDF \cite{mittal2022autosdf} & 18.18 & 12.13 & 19.17 & 16.49  \tabularnewline 
\multicolumn{1}{c|}{($\times 10^2$)}& \textit{Ours} & \textbf{6.996} & \textbf{6.599} & \textbf{10.87} & \textbf{8.155}  \tabularnewline 
\hline
MMD $\downarrow$ & AutoSDF \cite{mittal2022autosdf}  & 2.404 & 1.466  & 3.540 & 2.470  \tabularnewline 
\multicolumn{1}{c|}{($\times 10^3$)} & \textit{Ours} & \textbf{1.477} & \textbf{1.127} & \textbf{2.906} & \textbf{1.837}   \tabularnewline 
\hline
\end{tabular}}
\vspace{-0.08in}
\caption{Quantitative results of multi-modal partial point completion. The missing parts are always the top half.}
\label{tab:bottom_completion}
\vspace{-0.2in}
\end{table}

\subsection{Image-guide Generation \label{sec:image_generation}}
Next, we show the flexibility that ImAM can easily extend to image-guide generation, which is a more challenging task. The flexibility lies in that (1) it is implemented by the easiest way of feature concatenation; (2) the condition form of images is various, which could be 1-D feature vector or patch tokens, or 2-D feature maps. For simplicity, we use a pretrained CLIP model (\textit{i.e.}, ViT-B/32) to extract feature vectors of images as conditions. All experiments are conducted on ShapeNet dataset with rendered images.

\noindent \textbf{Baselines.} CLIP-Forge \cite{sanghi2022clip} is a flow-based model, which is trained with pairs of images and shapes. We take it as our primary baseline for two reasons: (1) it is a generation model instead of reconstruction model, and (2) it originally uses CLIP models to extract image features.

\noindent \textbf{Metrics and Settings.} We evaluate models with the test split of 13 categories. For each category, we randomly sample 50 singe-view images and then generate 5 shapes for evaluation. As a generation task, TMD is adopted to measure the diversity. We further use MMD with Chamfer Distance and Fr$\acute{\text{e}}$chet Point Cloud distance (FPD) \cite{shu20193d} to measure the fidelity compared with the ground truth.

\begin{figure}
 \begin{centering}
\includegraphics[scale=0.4]{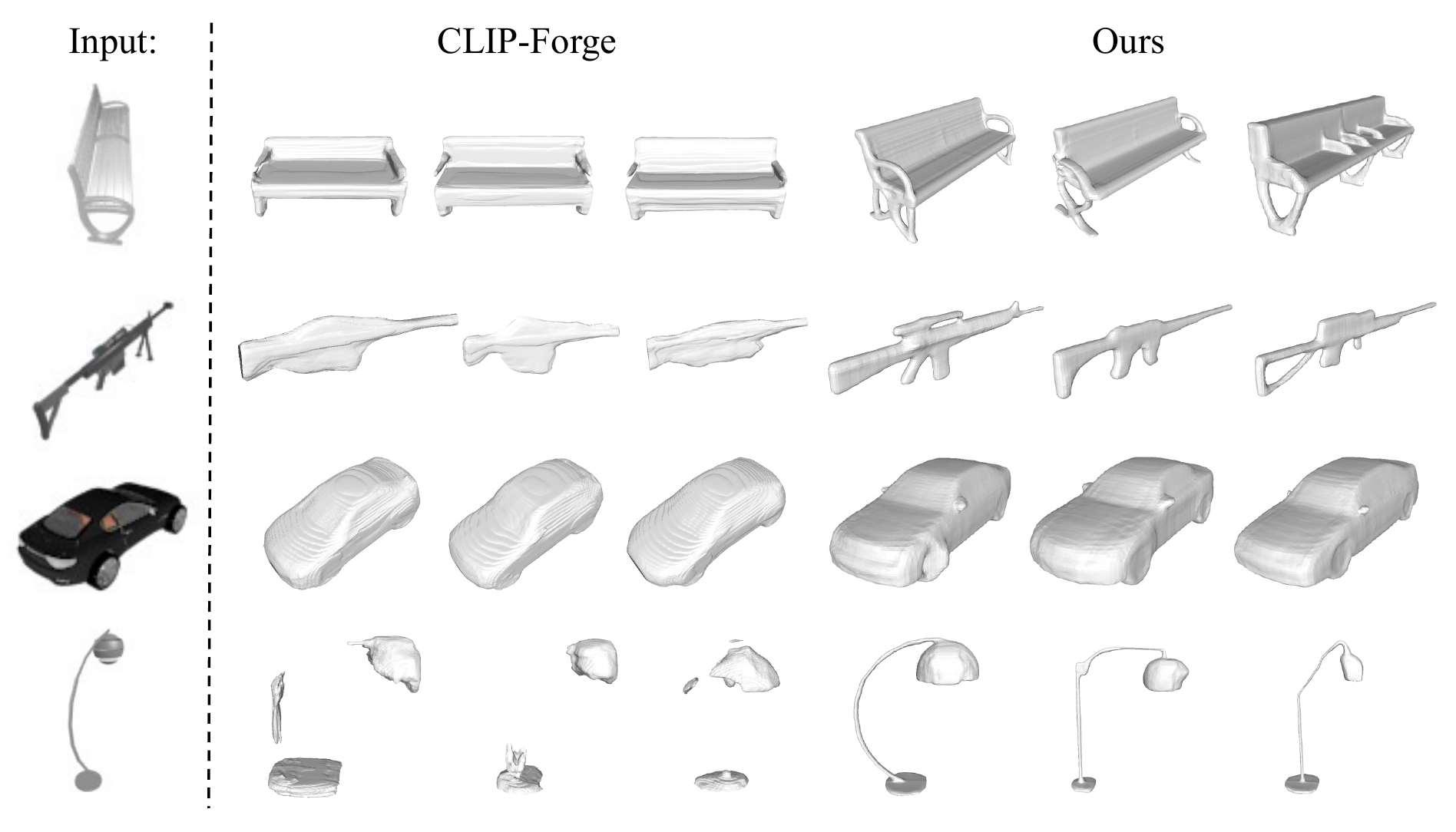}
\vspace{-0.1in}
\caption{Visualizations of image-guide shape generation. \label{fig:image_generation}}
\vspace{-0.2in}
\end{centering}
\end{figure}

\noindent \textbf{Results Analysis.} Results are reported in Tab.~\ref{tab:image_generation}. ImAM wins out in terms of both fidelity and diversity.
In particular, we achieve a great advantage on both TMD and FPD, demonstrating the effectiveness of ImAM applied to image-guide generation. It is also successfully echoed by qualitative visualizations in Fig.~\ref{fig:image_generation}. Our samples are diverse and appear visually faithful to attributes of the object in images.

\subsection{Text-guide Generation \label{sec:text_generation}}
Encouraged by promising results on image-guide generation, we also turn ImAM to text-to-shape generation. The same pretrained CLIP model is used to extract single embedding for text conditions. Note that we did it on purpose, not using word sequence embedding, but only using a single features vectors in our model to show its efficacy and scalability to the simplest forms of conditions.

\noindent \textbf{Data.} To our knowledge, the only existing largest paired text-shape dataset is Text2Shape \cite{chen2018text2shape}, which provides language descriptions for two objects from ShapeNet, \textit{i.e.}, chair and table. Thereby, we consider it as our main dataset to perform text-guide shape generation.

\noindent \textbf{Baselines.} We compare our model with two state-of-the-arts text-to-shape generation model. One is CLIP-Forge \cite{sanghi2022clip} under the supervised learning setting, using the same condition extractor as ours. The other is ITG \cite{liu2022towards}, which adopts BERT to encode texts into sequence embeddings.

\noindent \textbf{Metrics and Settings.} All models are trained on the train split of Text2Shape dataset with their official codes. We evaluate our model with two types of text queries: (1) description: the test split of Text2Shape; (2) prompts: customized short phrases provided by \cite{sanghi2022clip} containing attributes for chair and table. We additionally use Accuracy (Acc.) \cite{sanghi2022clip} to measure the fidelity. The Accuracy is calculated by a pretrained PointNet \cite{qi2017pointnet} classifier on ShapeNet.

\noindent \textbf{Results Analysis.} We report quantitative results of text-guide generation in Tab.~\ref{tab:text_generation}. ImAM achieves promising results on TMD, MMD and Acc, showing good generalization performance across different conditional generation tasks.

\begin{table}
\centering
\footnotesize
\setlength{\tabcolsep}{2.5mm}{
\begin{tabular}{lccc} 
\hline 
\multicolumn{1}{c}{\textsc{Method}} & TMD ($\times 10^2 $) $\uparrow$ & MMD ($\times 10^3$) $\downarrow$ & FPD $\downarrow$ \tabularnewline 
\hline 
AutoSDF \cite{mittal2022autosdf} & 2.657 & 2.137 & 15.262 \tabularnewline 
Clip-Forge \cite{sanghi2022clip} & 2.858 & 1.926 & 8.094   \tabularnewline 
\textit{Ours} & \textbf{4.274} & \textbf{1.590} & \textbf{1.680}  \tabularnewline 
\hline
\end{tabular}}
\vspace{-0.08in}
\caption{Quantitative results of image-guide generation. \label{tab:image_generation}}
\vspace{-0.1in}
\end{table}

\begin{table}
\centering
\footnotesize
\begin{subtable}{1.\linewidth}
	\centering
	\footnotesize
	\setlength{\tabcolsep}{2.3mm}{
	\begin{tabular}{lccc} 
	\hline 
	\multicolumn{1}{c}{\textsc{Method}} & TMD ($\times 10^1 $) $\uparrow$ & MMD ($\times 10^3$) $\downarrow$ & Acc $\uparrow$ \tabularnewline 
	\hline 
	ITG \cite{liu2022towards}  & N/A & 2.187  &  29.13  \tabularnewline 
        AutoSDF \cite{mittal2022autosdf} & 0.376 & 2.843 & 53.57 \tabularnewline 
	CLIP-Forge \cite{sanghi2022clip} & 0.400 & 2.136 &  53.68  \tabularnewline 
	\textit{Ours} & \textbf{0.565} & \textbf{1.846} & \textbf{59.93}  \tabularnewline 
	\hline
	\end{tabular}}
        \vspace{-0.05in}
	\caption{Descriptions as text queries \label{subtab:description}}
\end{subtable}
\begin{subtable}{1.\linewidth}
	\centering
	\setlength{\tabcolsep}{3.5mm}{
	\begin{tabular}{lccc} 
	\hline 
	\multicolumn{1}{c}{\textsc{Method}} & TMD ($\times 10^1 $) $\uparrow$ & FPD $\downarrow$ & Acc. $\uparrow$ \tabularnewline 
	\hline 
        AutoSDF \cite{mittal2022autosdf} & 0.842 & 29.47 & 48.70 \tabularnewline 
	CLIP-Forge \cite{sanghi2022clip} & 0.961 & \textbf{4.14} & 55.00  \tabularnewline 
	\textit{Ours} & \textbf{1.246} & 5.39 & \textbf{60.87}  \tabularnewline 
	\hline
	\end{tabular}}
        \vspace{-0.05in}
	\caption{Prompts as text queries \label{subtab:prompt}}
\end{subtable}
\vspace{-0.25in}
\caption{Quantitative results of text-guide generation. \label{tab:text_generation}}
\vspace{-0.2in}
\end{table}

\subsection{Ablation Study \label{sec:more_results}}

Lastly, we provide in-depth studies to dissect the efficacy of our ImAM framework, and several proposed designs. More discussions are presented in the Suppl. 

\noindent \textbf{Design Choices in Discrete Representation Learning.}
As a key contribution of this paper, we first discuss the efficacy of our improved discrete representation learning. We use `Vector' to denote our design since we apply vector quantization to latent vector. Similarly, 'Grid' and 'Tri-Plane' refer to baselines applying vector quantization to volumetric grids and the three planes, respectively. Results in Tab.~\ref{tab:ablation_study} show that `Grid' gets better IoU performance for shape reconstruction in the first stage, but fails to train transformers in the second stage due to extreme long length of sequence (\textit{e.g.}, $32^{3}$). 
In contrast, our model not only achieves comparable reconstruction results (\#0 \textit{vs.} \#2), but also outperforms 'Tri-Plane' by a large margin on generation quality (\#1 \textit{vs.} \#2). The latter shows inferior results due to `ambiguity of order' (see Suppl.)
It significantly proves the efficacy of our proposed coupling network, improving the plasticity and flexibility of the model.
We also explore different design choices in Tab.~\ref{tab:ablation_study}. By gradually increasing feature resolutions or the number of entries in the codebook, we achieve better performance on IoU. This observation is consistent with \cite{yu2021vector}, as the capacity of the discrete representation is affected by these two factors. We do not try larger parameters, as it would significantly increase the computation cost.

\begin{figure}
 \begin{centering}
\includegraphics[width=6.5cm,height=4cm]{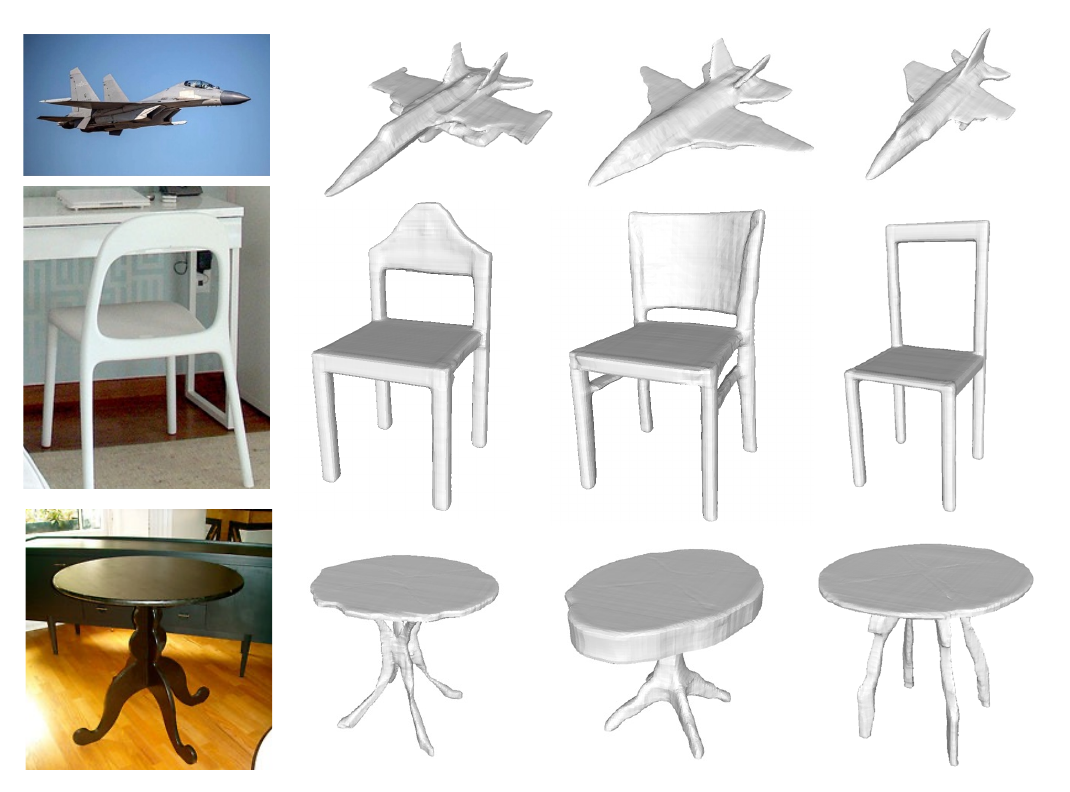}
\vspace{-0.1in}
\caption{Results of real-world image-guide generation. \label{fig:real_world}}
\vspace{-0.15in}
\end{centering}
\end{figure}

\noindent \textbf{Image-guide Generation in Real-world.} We further investigate the generalizability of our model on real-world images. We use the model trained on ShapeNet as described in Sec.~\ref{sec:image_generation}, and download images from internet as conditions. Figure~\ref{fig:real_world} shows the qualitative results for three categories, \textit{i.e.}, plane, chair and table. Our model sensitively capture major attributes of objects in the image and produce shapes faithful to them (see the first column to the left). Meanwhile, our synthesized samples enjoy the advantage of diversity by partially sticking to the images.

\begin{figure}
 \begin{centering}
\includegraphics[scale=0.38]{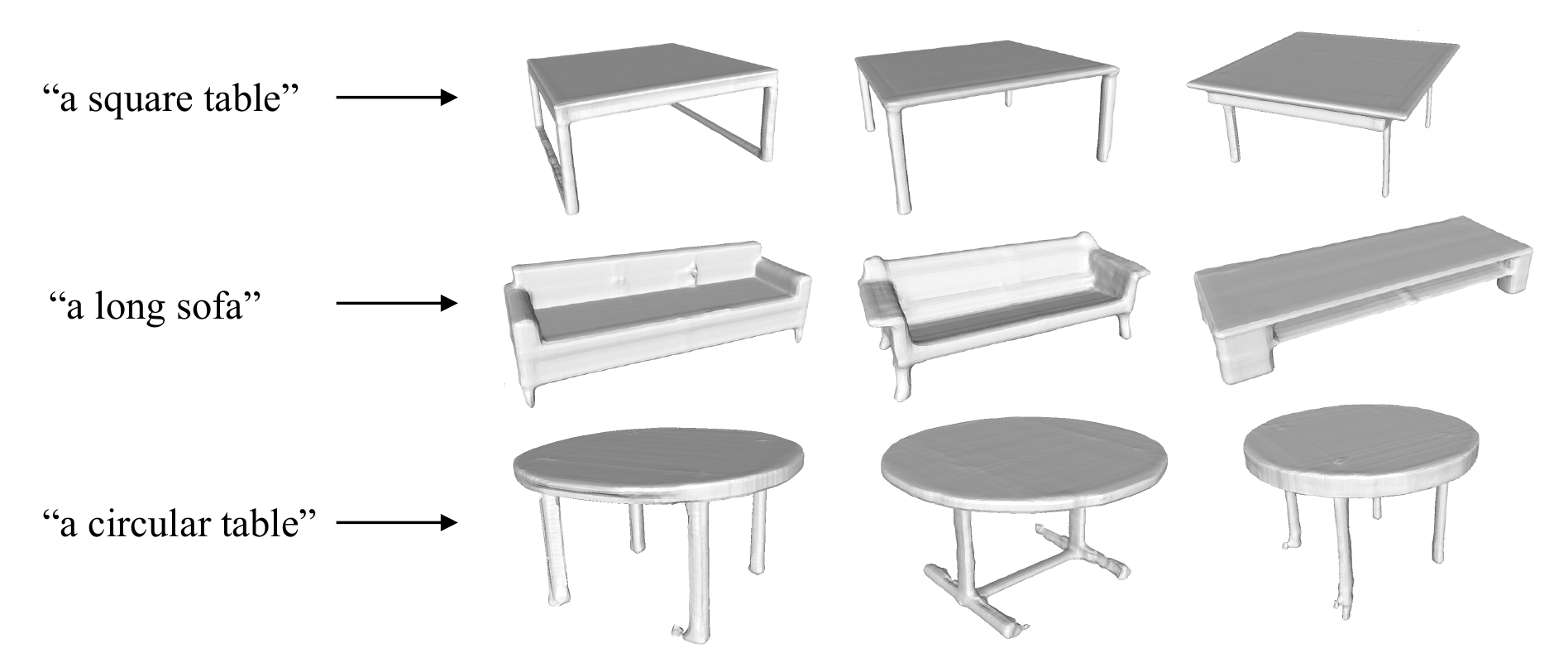}
\vspace{-0.1in}
\caption{Results of zero-shot text-to-shape generation. \label{fig:zero_shot}}
\vspace{-0.08in}
\end{centering}
\end{figure}

\noindent \textbf{Zero-shot Text-to-shape Generation.}
Inspired by CLIP-Forge \cite{sanghi2022clip}, we utilize the CLIP model to achieve zero-shot text-to-shape generation. At training, we only use the rendered images of 3D shapes. At inference, we substitute image features with text features encoded by the CLIP model. Figure~\ref{fig:zero_shot} shows our ability of zero-shot generation, 
where shape attributes are controlled with different prompts.

\begin{table}
\centering
\footnotesize 
\setlength{\tabcolsep}{1.6mm}{
\begin{tabular}{c|c|c|c|c|c} 
\hline
\multirow{2}{*}{\textsc{Num.}} &\multirow{2}{*}{\textsc{Type}} & \multirow{2}{*}{\textsc{\#Entry}} & \multirow{2}{*}{\textsc{Reso.}} & \textsc{Stage 1} & \textsc{Stage 2} \tabularnewline 
\cline{5-6}
& & & & IoU $\uparrow$ & 1-NNA / ECD $\downarrow$  \tabularnewline 
\hline
\hline 
0 & Grid & \multirow{3}{*}{4096} & \multirow{3}{*}{32} & 88.87  & $\times$  \tabularnewline 
\cline{2-2} \cline{5-6}
1 & Tri-Plane &  & & 87.81 & 73.67 / 743 \tabularnewline 
\cline{2-2} \cline{5-6}
2 & \multirow{4}{*}{Vector} &  & & 88.01 & \bf{59.95 / 236} \tabularnewline 
\cline{3-6}
3 &  & 4096 & 16 & 79.17 & \multirow{3}{*}{-} \tabularnewline 
\cline{3-5}
4 &  & 2048 & \multirow{2}{*}{32} & 86.99 &  \tabularnewline 
\cline{3-3}\cline{5-5}
5 &  & 1024 &  & 86.57 &  \tabularnewline 
\hline
\end{tabular}}
\vspace{-0.08in}
\caption{Ablation study of auto-encoder design choices. We report 1-NNA/ECD for plane category. `\textsc{Reso.}' means the resolution of feature map for vector quantization. `$\times$': cannot report due to extreme memory cost. `-': not report. 
\bf{\textit{Notably, without the coupling network, our method naturally degenerates into `Tri-Plane' representation.}}
\label{tab:ablation_study}}
\vspace{-0.2in}
\end{table}

\section{Conclusion \label{sec:conclusion}}
We introduce an improved AR model for 3D shape generation. By projecting volumetric grids of encoded input shapes onto three axis-aligned orthogonal feature planes, which are then coupled into a latent vector, we reduce computational costs and create a more tractable order for AR learning. Our compact and tractable representations enable easy switching between unconditional and conditional generation with multi-modal conditioning inputs. Extensive experiments show that our model outperforms previous methods on multiple generation tasks.

{\small
\bibliographystyle{ieee_fullname}
\bibliography{egbib}

\begin{thebibliography}{10}\itemsep=-1pt

\bibitem{achlioptas2018learning}
Panos Achlioptas, Olga Diamanti, Ioannis Mitliagkas, and Leonidas Guibas.
\newblock Learning representations and generative models for 3d point clouds.
\newblock In {\em International conference on machine learning}, pages 40--49.
  PMLR, 2018.

\bibitem{cai2020learning}
Ruojin Cai, Guandao Yang, Hadar Averbuch-Elor, Zekun Hao, Serge Belongie, Noah
  Snavely, and Bharath Hariharan.
\newblock Learning gradient fields for shape generation.
\newblock In {\em European Conference on Computer Vision}, pages 364--381.
  Springer, 2020.

\bibitem{chan2021efficient}
Eric~R Chan, Connor~Z Lin, Matthew~A Chan, Koki Nagano, Boxiao Pan, Shalini
  De~Mello, Orazio Gallo, Leonidas Guibas, Jonathan Tremblay, Sameh Khamis,
  et~al.
\newblock Efficient geometry-aware 3d generative adversarial networks.
\newblock {\em arXiv preprint arXiv:2112.07945}, 2021.

\bibitem{chang2015shapenet}
Angel~X Chang, Thomas Funkhouser, Leonidas Guibas, Pat Hanrahan, Qixing Huang,
  Zimo Li, Silvio Savarese, Manolis Savva, Shuran Song, Hao Su, et~al.
\newblock Shapenet: An information-rich 3d model repository.
\newblock {\em arXiv preprint arXiv:1512.03012}, 2015.

\bibitem{chang2022maskgit}
Huiwen Chang, Han Zhang, Lu Jiang, Ce Liu, and William~T Freeman.
\newblock Maskgit: Masked generative image transformer.
\newblock {\em arXiv preprint arXiv:2202.04200}, 2022.

\bibitem{chen2003visual}
Ding-Yun Chen, Xiao-Pei Tian, Yu-Te Shen, and Ming Ouhyoung.
\newblock On visual similarity based 3d model retrieval.
\newblock In {\em Computer graphics forum}, volume~22, pages 223--232. Wiley
  Online Library, 2003.

\bibitem{chen2018text2shape}
Kevin Chen, Christopher~B Choy, Manolis Savva, Angel~X Chang, Thomas
  Funkhouser, and Silvio Savarese.
\newblock Text2shape: Generating shapes from natural language by learning joint
  embeddings.
\newblock In {\em Asian conference on computer vision}, pages 100--116.
  Springer, 2018.

\bibitem{chen2019learning}
Zhiqin Chen and Hao Zhang.
\newblock Learning implicit fields for generative shape modeling.
\newblock In {\em Proceedings of the IEEE/CVF Conference on Computer Vision and
  Pattern Recognition}, pages 5939--5948, 2019.

\bibitem{cheng2022autoregressive}
An-Chieh Cheng, Xueting Li, Sifei Liu, Min Sun, and Ming-Hsuan Yang.
\newblock Autoregressive 3d shape generation via canonical mapping.
\newblock {\em arXiv preprint arXiv:2204.01955}, 2022.

\bibitem{choy20163d}
Christopher~B Choy, Danfei Xu, JunYoung Gwak, Kevin Chen, and Silvio Savarese.
\newblock 3d-r2n2: A unified approach for single and multi-view 3d object
  reconstruction.
\newblock In {\em European conference on computer vision}, pages 628--644.
  Springer, 2016.

\bibitem{esser2021taming}
Patrick Esser, Robin Rombach, and Bjorn Ommer.
\newblock Taming transformers for high-resolution image synthesis.
\newblock In {\em Proceedings of the IEEE/CVF Conference on Computer Vision and
  Pattern Recognition}, pages 12873--12883, 2021.

\bibitem{fan2017point}
Haoqiang Fan, Hao Su, and Leonidas~J Guibas.
\newblock A point set generation network for 3d object reconstruction from a
  single image.
\newblock In {\em Proceedings of the IEEE conference on computer vision and
  pattern recognition}, pages 605--613, 2017.

\bibitem{hane2017hierarchical}
Christian H{\"a}ne, Shubham Tulsiani, and Jitendra Malik.
\newblock Hierarchical surface prediction for 3d object reconstruction.
\newblock In {\em 2017 International Conference on 3D Vision (3DV)}, pages
  412--420. IEEE, 2017.

\bibitem{ibing20213d}
Moritz Ibing, Isaak Lim, and Leif Kobbelt.
\newblock 3d shape generation with grid-based implicit functions.
\newblock In {\em Proceedings of the IEEE/CVF Conference on Computer Vision and
  Pattern Recognition}, pages 13559--13568, 2021.

\bibitem{jiang2020local}
Chiyu Jiang, Avneesh Sud, Ameesh Makadia, Jingwei Huang, Matthias Nie{\ss}ner,
  Thomas Funkhouser, et~al.
\newblock Local implicit grid representations for 3d scenes.
\newblock In {\em Proceedings of the IEEE/CVF Conference on Computer Vision and
  Pattern Recognition}, pages 6001--6010, 2020.

\bibitem{kazhdan2006poisson}
Michael Kazhdan, Matthew Bolitho, and Hugues Hoppe.
\newblock Poisson surface reconstruction.
\newblock In {\em Proceedings of the fourth Eurographics symposium on Geometry
  processing}, volume~7, 2006.

\bibitem{liu2022towards}
Zhengzhe Liu, Yi Wang, Xiaojuan Qi, and Chi-Wing Fu.
\newblock Towards implicit text-guided 3d shape generation.
\newblock In {\em Proceedings of the IEEE/CVF Conference on Computer Vision and
  Pattern Recognition}, pages 17896--17906, 2022.

\bibitem{lorensen1987marching}
William~E Lorensen and Harvey~E Cline.
\newblock Marching cubes: A high resolution 3d surface construction algorithm.
\newblock {\em ACM siggraph computer graphics}, 21(4):163--169, 1987.

\bibitem{maturana2015voxnet}
Daniel Maturana and Sebastian Scherer.
\newblock Voxnet: A 3d convolutional neural network for real-time object
  recognition.
\newblock In {\em 2015 IEEE/RSJ International Conference on Intelligent Robots
  and Systems (IROS)}, pages 922--928. IEEE, 2015.

\bibitem{mees2019self}
Oier Mees, Maxim Tatarchenko, Thomas Brox, and Wolfram Burgard.
\newblock Self-supervised 3d shape and viewpoint estimation from single images
  for robotics.
\newblock In {\em 2019 IEEE/RSJ International Conference on Intelligent Robots
  and Systems (IROS)}, pages 6083--6089. IEEE, 2019.

\bibitem{mescheder2019occupancy}
Lars Mescheder, Michael Oechsle, Michael Niemeyer, Sebastian Nowozin, and
  Andreas Geiger.
\newblock Occupancy networks: Learning 3d reconstruction in function space.
\newblock In {\em Proceedings of the IEEE/CVF Conference on Computer Vision and
  Pattern Recognition}, pages 4460--4470, 2019.

\bibitem{michalkiewicz2019implicit}
Mateusz Michalkiewicz, Jhony~K Pontes, Dominic Jack, Mahsa Baktashmotlagh, and
  Anders Eriksson.
\newblock Implicit surface representations as layers in neural networks.
\newblock In {\em Proceedings of the IEEE/CVF International Conference on
  Computer Vision}, pages 4743--4752, 2019.

\bibitem{mittal2022autosdf}
Paritosh Mittal, Yen-Chi Cheng, Maneesh Singh, and Shubham Tulsiani.
\newblock Autosdf: Shape priors for 3d completion, reconstruction and
  generation.
\newblock In {\em Proceedings of the IEEE/CVF Conference on Computer Vision and
  Pattern Recognition}, pages 306--315, 2022.

\bibitem{nash2020polygen}
Charlie Nash, Yaroslav Ganin, SM~Ali Eslami, and Peter Battaglia.
\newblock Polygen: An autoregressive generative model of 3d meshes.
\newblock In {\em International Conference on Machine Learning}, pages
  7220--7229. PMLR, 2020.

\bibitem{park2019deepsdf}
Jeong~Joon Park, Peter Florence, Julian Straub, Richard Newcombe, and Steven
  Lovegrove.
\newblock Deepsdf: Learning continuous signed distance functions for shape
  representation.
\newblock In {\em Proceedings of the IEEE/CVF Conference on Computer Vision and
  Pattern Recognition}, pages 165--174, 2019.

\bibitem{peng2020convolutional}
Songyou Peng, Michael Niemeyer, Lars Mescheder, Marc Pollefeys, and Andreas
  Geiger.
\newblock Convolutional occupancy networks.
\newblock In {\em European Conference on Computer Vision}, pages 523--540.
  Springer, 2020.

\bibitem{qi2017pointnet}
Charles~R Qi, Hao Su, Kaichun Mo, and Leonidas~J Guibas.
\newblock Pointnet: Deep learning on point sets for 3d classification and
  segmentation.
\newblock In {\em Proceedings of the IEEE conference on computer vision and
  pattern recognition}, pages 652--660, 2017.

\bibitem{qian2022impdet}
Xuelin Qian, Li Wang, Yi Zhu, Li Zhang, Yanwei Fu, and Xiangyang Xue.
\newblock Impdet: Exploring implicit fields for 3d object detection.
\newblock {\em arXiv preprint arXiv:2203.17240}, 2022.

\bibitem{razavi2019generating}
Ali Razavi, Aaron Van~den Oord, and Oriol Vinyals.
\newblock Generating diverse high-fidelity images with vq-vae-2.
\newblock {\em Advances in neural information processing systems}, 32, 2019.

\bibitem{riegler2017octnetfusion}
Gernot Riegler, Ali~Osman Ulusoy, Horst Bischof, and Andreas Geiger.
\newblock Octnetfusion: Learning depth fusion from data.
\newblock In {\em 2017 International Conference on 3D Vision (3DV)}, pages
  57--66. IEEE, 2017.

\bibitem{sanghi2022clip}
Aditya Sanghi, Hang Chu, Joseph~G Lambourne, Ye Wang, Chin-Yi Cheng, Marco
  Fumero, and Kamal~Rahimi Malekshan.
\newblock Clip-forge: Towards zero-shot text-to-shape generation.
\newblock In {\em Proceedings of the IEEE/CVF Conference on Computer Vision and
  Pattern Recognition}, pages 18603--18613, 2022.

\bibitem{shu20193d}
Dong~Wook Shu, Sung~Woo Park, and Junseok Kwon.
\newblock 3d point cloud generative adversarial network based on tree
  structured graph convolutions.
\newblock In {\em Proceedings of the IEEE/CVF international conference on
  computer vision}, pages 3859--3868, 2019.

\bibitem{sitzmann2020metasdf}
Vincent Sitzmann, Eric Chan, Richard Tucker, Noah Snavely, and Gordon
  Wetzstein.
\newblock Metasdf: Meta-learning signed distance functions.
\newblock {\em Advances in Neural Information Processing Systems},
  33:10136--10147, 2020.

\bibitem{stets2017visualization}
Jonathan~Dyssel Stets, Yongbin Sun, Wiley Corning, and Scott~W Greenwald.
\newblock Visualization and labeling of point clouds in virtual reality.
\newblock In {\em SIGGRAPH Asia 2017 Posters}, pages 1--2. 2017.

\bibitem{sun2018x}
Yongbin Sun, Sai Nithin~R Kantareddy, Rahul Bhattacharyya, and Sanjay~E Sarma.
\newblock X-vision: An augmented vision tool with real-time sensing ability in
  tagged environments.
\newblock In {\em 2018 ieee international conference on rfid technology \&
  application (rfid-ta)}, pages 1--6. IEEE, 2018.

\bibitem{sun2020pointgrow}
Yongbin Sun, Yue Wang, Ziwei Liu, Joshua Siegel, and Sanjay Sarma.
\newblock Pointgrow: Autoregressively learned point cloud generation with
  self-attention.
\newblock In {\em Proceedings of the IEEE/CVF Winter Conference on Applications
  of Computer Vision}, pages 61--70, 2020.

\bibitem{van2016conditional}
Aaron Van~den Oord, Nal Kalchbrenner, Lasse Espeholt, Oriol Vinyals, Alex
  Graves, et~al.
\newblock Conditional image generation with pixelcnn decoders.
\newblock {\em Advances in neural information processing systems}, 29, 2016.

\bibitem{van2017neural}
Aaron Van Den~Oord, Oriol Vinyals, et~al.
\newblock Neural discrete representation learning.
\newblock {\em Advances in neural information processing systems}, 30, 2017.

\bibitem{vaswani2017attention}
Ashish Vaswani, Noam Shazeer, Niki Parmar, Jakob Uszkoreit, Llion Jones,
  Aidan~N Gomez, {\L}ukasz Kaiser, and Illia Polosukhin.
\newblock Attention is all you need.
\newblock {\em Advances in Neural Information Processing Systems}, 30, 2017.

\bibitem{wang2017cnn}
Peng-Shuai Wang, Yang Liu, Yu-Xiao Guo, Chun-Yu Sun, and Xin Tong.
\newblock O-cnn: Octree-based convolutional neural networks for 3d shape
  analysis.
\newblock {\em ACM Transactions On Graphics (TOG)}, 36(4):1--11, 2017.

\bibitem{wu2016learning}
Jiajun Wu, Chengkai Zhang, Tianfan Xue, Bill Freeman, and Josh Tenenbaum.
\newblock Learning a probabilistic latent space of object shapes via 3d
  generative-adversarial modeling.
\newblock {\em Advances in neural information processing systems}, 29, 2016.

\bibitem{wu2020multimodal}
Rundi Wu, Xuelin Chen, Yixin Zhuang, and Baoquan Chen.
\newblock Multimodal shape completion via conditional generative adversarial
  networks.
\newblock In {\em European Conference on Computer Vision}, pages 281--296.
  Springer, 2020.

\bibitem{wu2019pointconv}
Wenxuan Wu, Zhongang Qi, and Li Fuxin.
\newblock Pointconv: Deep convolutional networks on 3d point clouds.
\newblock In {\em Proceedings of the IEEE/CVF Conference on Computer Vision and
  Pattern Recognition}, pages 9621--9630, 2019.

\bibitem{wu20153d}
Zhirong Wu, Shuran Song, Aditya Khosla, Fisher Yu, Linguang Zhang, Xiaoou Tang,
  and Jianxiong Xiao.
\newblock 3d shapenets: A deep representation for volumetric shapes.
\newblock In {\em Proceedings of the IEEE conference on computer vision and
  pattern recognition}, pages 1912--1920, 2015.

\bibitem{yan2022shapeformer}
Xingguang Yan, Liqiang Lin, Niloy~J Mitra, Dani Lischinski, Daniel Cohen-Or,
  and Hui Huang.
\newblock Shapeformer: Transformer-based shape completion via sparse
  representation.
\newblock In {\em Proceedings of the IEEE/CVF Conference on Computer Vision and
  Pattern Recognition}, pages 6239--6249, 2022.

\bibitem{yang2019pointflow}
Guandao Yang, Xun Huang, Zekun Hao, Ming-Yu Liu, Serge Belongie, and Bharath
  Hariharan.
\newblock Pointflow: 3d point cloud generation with continuous normalizing
  flows.
\newblock In {\em Proceedings of the IEEE/CVF International Conference on
  Computer Vision}, pages 4541--4550, 2019.

\bibitem{ye2021online}
Jianglong Ye, Yuntao Chen, Naiyan Wang, and Xiaolong Wang.
\newblock Online adaptation for implicit object tracking and shape
  reconstruction in the wild.
\newblock {\em arXiv preprint arXiv:2111.12728}, 2021.

\bibitem{yu2021vector}
Jiahui Yu, Xin Li, Jing~Yu Koh, Han Zhang, Ruoming Pang, James Qin, Alexander
  Ku, Yuanzhong Xu, Jason Baldridge, and Yonghui Wu.
\newblock Vector-quantized image modeling with improved vqgan.
\newblock {\em arXiv preprint arXiv:2110.04627}, 2021.

\bibitem{yu2021pointr}
Xumin Yu, Yongming Rao, Ziyi Wang, Zuyan Liu, Jiwen Lu, and Jie Zhou.
\newblock Pointr: Diverse point cloud completion with geometry-aware
  transformers.
\newblock In {\em Proceedings of the IEEE/CVF International Conference on
  Computer Vision}, pages 12498--12507, 2021.

\bibitem{zhao2021improved}
Long Zhao, Zizhao Zhang, Ting Chen, Dimitris Metaxas, and Han Zhang.
\newblock Improved transformer for high-resolution gans.
\newblock {\em Advances in Neural Information Processing Systems}, 34, 2021.

\bibitem{zhou20213d}
Linqi Zhou, Yilun Du, and Jiajun Wu.
\newblock 3d shape generation and completion through point-voxel diffusion.
\newblock In {\em Proceedings of the IEEE/CVF International Conference on
  Computer Vision}, pages 5826--5835, 2021.

\end{thebibliography}
}

\newpage
\section*{\Large{Appendix}}
\section*{A. Motivation of ImAM}
\subsection*{A.1. What Is the ‘Ambiguity’ in AR Models?\label{sec:ambiguity}}

Formally, \textit{`ambiguity' appears in the order of a series of conditional probabilities, which affects the difficulty of likelihood learning, leading to approximation error of the joint distribution.} 
Critically, in the second stage of AR models, 
it requires sequential outputs, autoregressively predicting the next code conditioned on all previous ones. Thereby, the order of the flattened sequence determines the order of conditional probabilities (Eq.~\textcolor{red}{5} and \textcolor{red}{6} in the main paper).
Although some methods (\eg position embedding \cite{vaswani2017attention}) can be aware of positions of codes, it cannot eliminate approximation error caused by the condition order. Notably, this `ambiguity' phenomenon is also discussed in \cite{esser2021taming} (Fig.~\textcolor{red}{47}).
Figure~\ref{fig:flatten_order} illustrates how the flattening order affects the way of autoregressive generation.
For grid-based representation, it is ambiguous if the flattening order along axes is $x$-$y$-$z$, $z$-$x$-$y$ or other combinations. Similarly, the flattening order for tri-planar representation is ambiguous, \eg, $p\left(\mathbf{z}^{xz}\right)p\left(\mathbf{z}^{xy}|\mathbf{z}^{xz}\right)p\left(\mathbf{z}^{yz}|\mathbf{z}^{xz},\mathbf{z}^{xy}\right)$, $p\left(\mathbf{z}^{yz}\right)p\left(\mathbf{z}^{xz}|\mathbf{z}^{yz}\right)p\left(\mathbf{z}^{xy}|\mathbf{z}^{yz},\mathbf{z}^{xz}\right)$ or others.

\subsection*{A.2. Effect of Flattening Orders \label{sec:effec_order}}

\begin{figure}[t]
\begin{centering}
\includegraphics[scale=0.36]{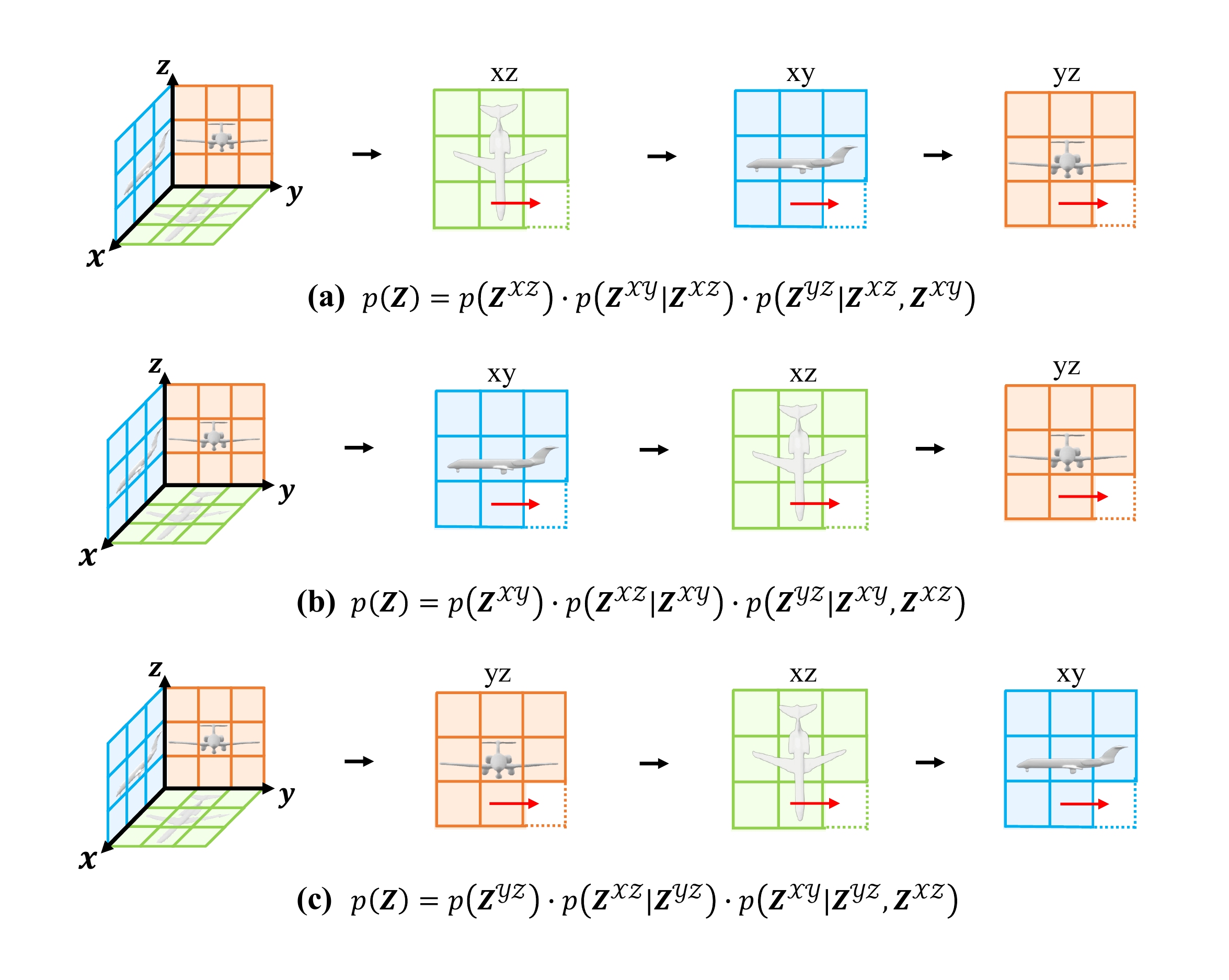}  
\tabularnewline
\vspace{-0.2in}
\caption{Illustration of auto-regressive generation for tri-planar representation. Here, we show three different flattening orders as examples.
\label{fig:flatten_order}}
\vspace{-0.1in}
\end{centering}
\end{figure}

We first investigate how the flattening order affects the quality of shape generation. This study takes tri-planar representation as an example (`Tri-Plane' for short), since \textit{our improved discrete representation (`Vector') can naturally degenerate into `Tri-Plane' by removing the proposed coupling network}. As illustrate in Fig.~\ref{fig:flatten_order}, different flattening order will affect different auto-regressive generation order of the three planes. Quantitatively, we consider three variants to learn joint distributions of tri-planar representation without loss of generality, Iter-A: $p\left(\mathbf{z}\right)=p\left(\mathbf{z}^{xz}\right)\cdot p\left(\mathbf{z}^{xy}|\mathbf{z}^{xz}\right)\cdot p\left(\mathbf{z}^{yz}|\mathbf{z}^{xz},\mathbf{z}^{xy}\right)$, Iter-B: $p\left(\mathbf{z}\right)=p\left(\mathbf{z}^{xy}\right)\cdot p\left(\mathbf{z}^{xz}|\mathbf{z}^{xy}\right)\cdot p\left(\mathbf{z}^{yz}|\mathbf{z}^{xy},\mathbf{z}^{xz}\right)$ and Iter-C:
$p\left(\mathbf{z}\right)=p\left(\mathbf{z}^{yz}\right)\cdot p\left(\mathbf{z}^{xz}|\mathbf{z}^{yz}\right)\cdot p\left(\mathbf{z}^{xy}|\mathbf{z}^{yz},\mathbf{z}^{xz}\right)$.

\begin{figure}[t]
\hspace{-0.14in}
\includegraphics[scale=0.26]{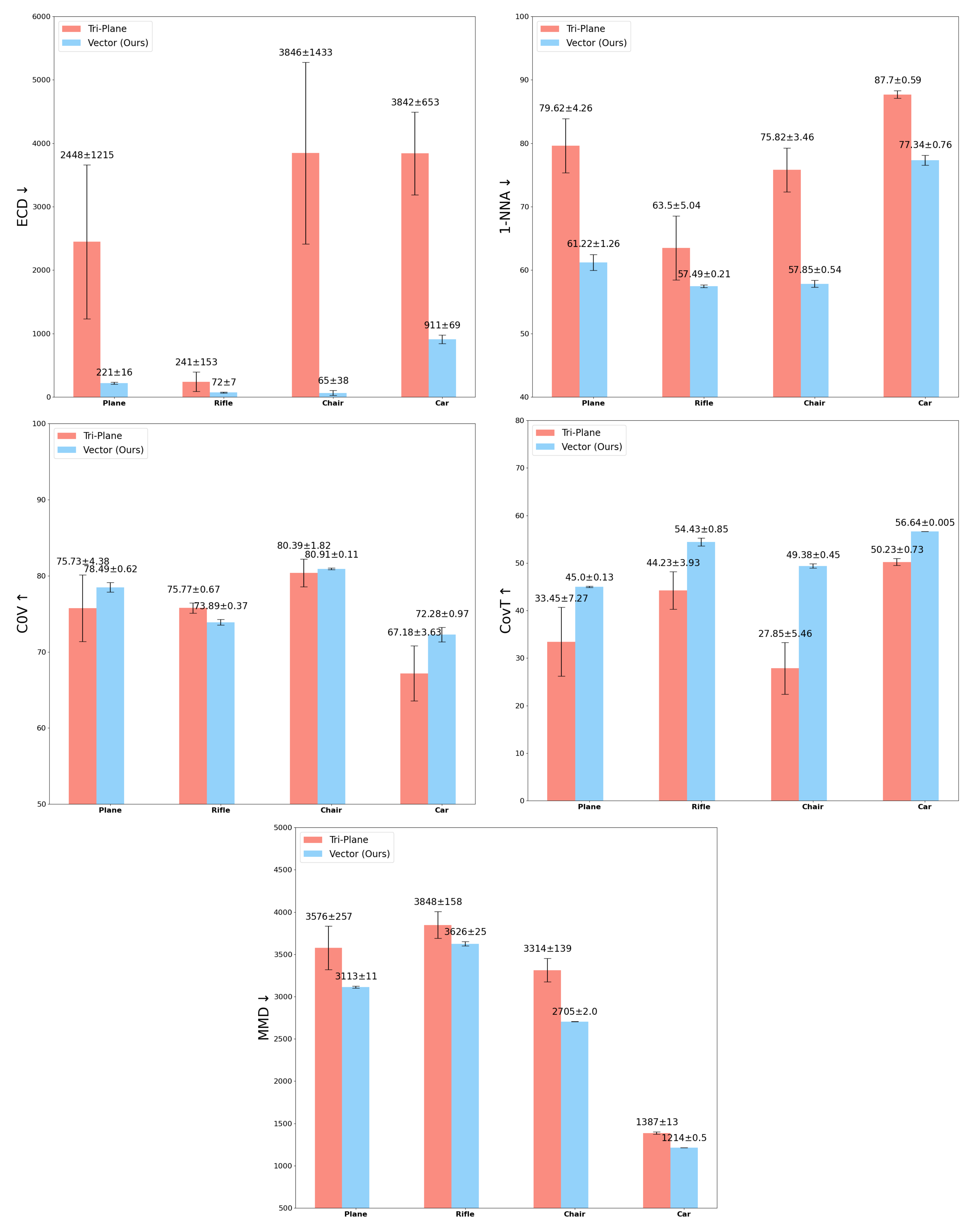}  
\tabularnewline
\vspace{-0.15in}
\caption{Statistic analysis of the effect of flattening order. We report mean and standard deviation as histogram and error bar, respectively.
Best viewed in color and zoom in.
\label{fig:order_hist}}
\vspace{-0.1in}
\end{figure}

\begin{figure}[t]
\begin{centering}
\includegraphics[scale=0.45]{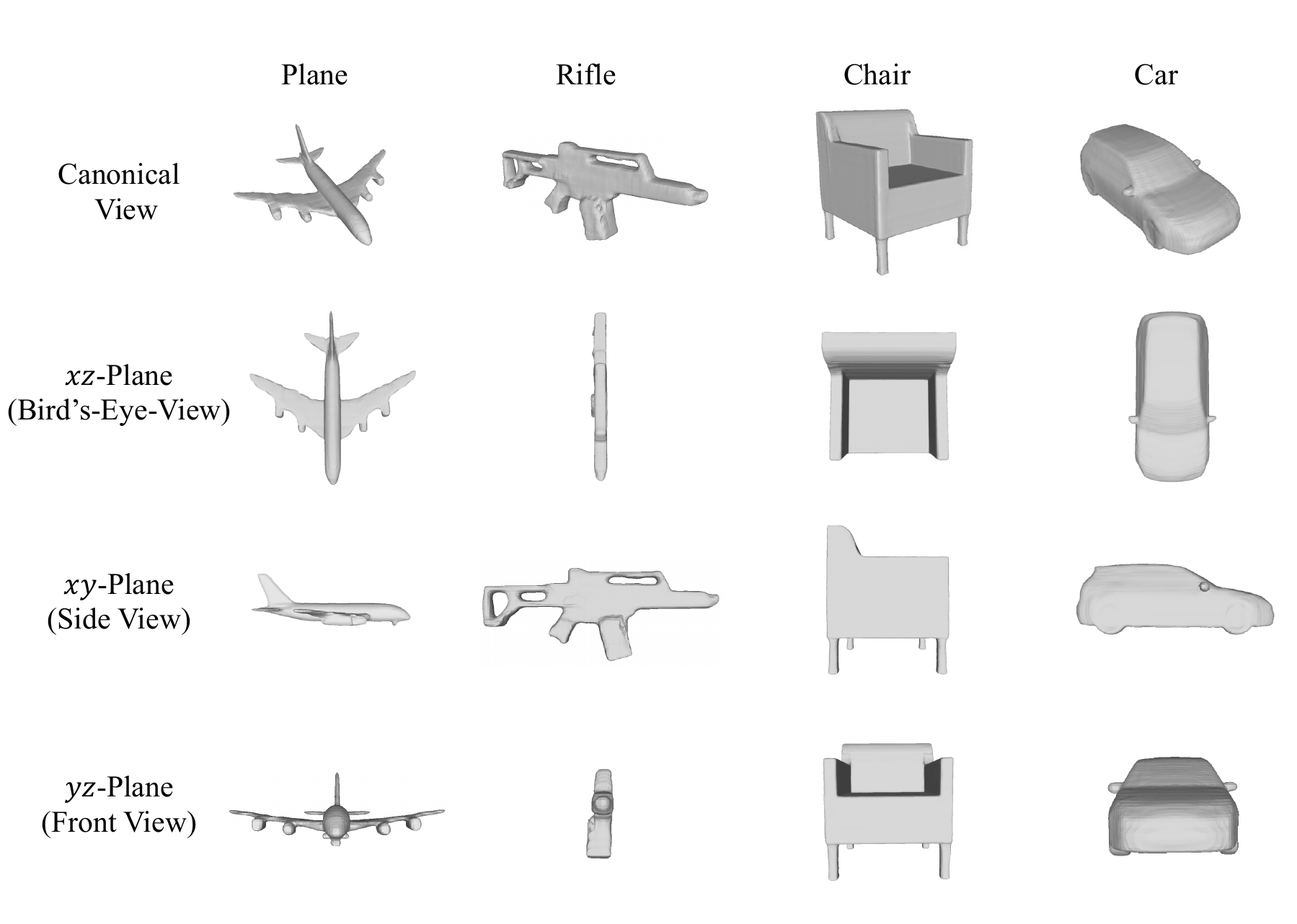}  
\tabularnewline
\vspace{-0.08in}
\caption{Visualizations of projected shapes on three planes.
\label{fig:diff_view}}
\vspace{-0.2in}
\end{centering}
\end{figure}

Results are presented in Tab.~\ref{tab:uncond} and Fig.~\ref{fig:order_hist}. As observed, different orders have significant impact on performance, resulting in a large value of standard deviation. For instance, Iter-A achieves a better result on Plane category (Iter-A: 73.67 \textit{vs.} Iter-B: 83.37 on 1-NNA), while for Rifle, it prefers the order of Iter-B (Iter-B: 56.54  \textit{vs.} Iter-A: 68.35 on 1-NNA).
We attempt to explain this phenomenon by visualizing projected shapes on three planes. As illustrated in Fig.~\ref{fig:diff_view}, for Plane category (First Column), the projection onto the $xy$-plane provides limited shape information, while the $xz$-plane reveals more representative geometries of the aircraft.
We agree that if the $xz$-plane that contains more shape information is generated first, the generation of subsequent planes may be much easier.
Consequently, it is beneficial for Iter-A to generate more faithful 3D shapes than Iter-B. In contrast, Rifle and Chair exhibit more details on $xy$-plane, so the autoregressive order of Iter-B yields better results for these two categories. In addition, we notice that Car has a relatively simple shape, \eg, a cuboid, leading to similar impacts on the generation quality for different flattening orders.

\begin{table}[htp]
\centering
\footnotesize 
\setlength{\tabcolsep}{1.5mm}{
\begin{tabular}{c|c|l|c|c|c|c}
\hline
\multirow{2}{*}{\textsc{Metrics}} &  \multicolumn{2}{c|}{\multirow{2}{*}{\textsc{Methods}}} & \multicolumn{4}{c}{\textsc{Categories}}  \tabularnewline
\cline{4-7}
 & \multicolumn{1}{c}{} & & Plane & Rifle & Chair & Car \tabularnewline 
\hline
\hline
\multirow{5}{*}{ECD $\downarrow$} & \multirow{3}{*}{Tri-Plane}
& Iter-A & 744 & 405 & 4966 & 3599 \tabularnewline
& & Iter-B & 3501 & \textbf{36} & 1823 & 4735 \tabularnewline
& & Iter-C & 3098 & 282 & 4749 & 3193 \tabularnewline
\cline{2-7}
& \multirow{2}{*}{Vector (\textit{ours})} & Row-Major & 236  & 65 & \textbf{27} & \textbf{842} \tabularnewline
& & Col-Major & \textbf{205} & 79 & 102 & 980 \tabularnewline

\hline
\multirow{5}{*}{1-NNA $\downarrow$} & \multirow{3}{*}{Tri-Plane}
& Iter-A & 73.67 & 68.35 & 78.15 & 87.16\tabularnewline
& & Iter-B & 83.37 & \textbf{56.54} & 70.92 & 87.42 \tabularnewline
& & Iter-C & 81.83 & 65.61 & 78.38 & 88.53 \tabularnewline
\cline{2-7}
& \multirow{2}{*}{Vector (\textit{ours})} & Row-Major & \textbf{59.95}  & 57.28 & \textbf{57.31} & \textbf{76.58} \tabularnewline
& & Col-Major & 62.48 & 57.70 & 58.38  & 78.09 \tabularnewline

\hline
\multirow{5}{*}{COV $\uparrow$} & \multirow{3}{*}{Tri-Plane}
& Iter-A & \textbf{81.70} & 75.10 & 79.33 & 65.31 \tabularnewline
& & Iter-B & 74.16 & 75.52 & \textbf{82.95} & 63.97\tabularnewline
& & Iter-C & 71.32 & \textbf{76.69} & 78.89 & 72.25\tabularnewline
\cline{2-7}
& \multirow{2}{*}{Vector (\textit{ours})} & Row-Major & 79.11 & 74.26 & 80.81 & \textbf{73.25} \tabularnewline
&  & Col-Major & 77.87 & 73.52 & 81.03 & 71.31\tabularnewline

\hline
\multirow{5}{*}{CovT $\uparrow$} & \multirow{3}{*}{Tri-Plane}
& Iter-A & 43.51 & 41.56 & 23.10 & 50.30 \tabularnewline
& & Iter-B & 26.57 & 49.78 & 35.50 & 49.43 \tabularnewline
& & Iter-C & 30.28 & 41.35 & 24.94 & 51.23 \tabularnewline
\cline{2-7}
& \multirow{2}{*}{Vector (\textit{ours})} & Row-Major & \textbf{45.12} & \textbf{55.27} & \textbf{49.82} & \textbf{56.64} \tabularnewline
& & Col-Major & 44.87 & 53.58 & 48.93 & 56.63 \tabularnewline

\hline
\multirow{5}{*}{MMD $\downarrow$} & \multirow{3}{*}{Tri-Plane}
& Iter-A & 3237 & 3962 & 3392 & 1373 \tabularnewline
& & Iter-B & 3860 & 3624 & 3119 & 1404 \tabularnewline
& & Iter-C & 3631 & 3958 & 3430 & 1385 \tabularnewline
\cline{2-7}
& \multirow{2}{*}{Vector (\textit{ours})} & Row-Major & 3124 & 3628  & \textbf{2703} & \textbf{1213} \tabularnewline
& & Col-Major & \textbf{3102} & \textbf{3623} & 2707 & 1214 \tabularnewline
\hline
\end{tabular}}
\vspace{-0.08in}
\caption{The effect of flattening order on different discrete representations. Here, we take unconditional generation as an example and train one model per class. 
Please find statistic and qualitative analyses in Fig.~\ref{fig:order_hist} and \ref{fig:diff_view}, respectively.
\label{tab:uncond}}
\vspace{-0.1in}
\end{table}

\begin{figure}[t]
\begin{centering}
\includegraphics[scale=0.38]{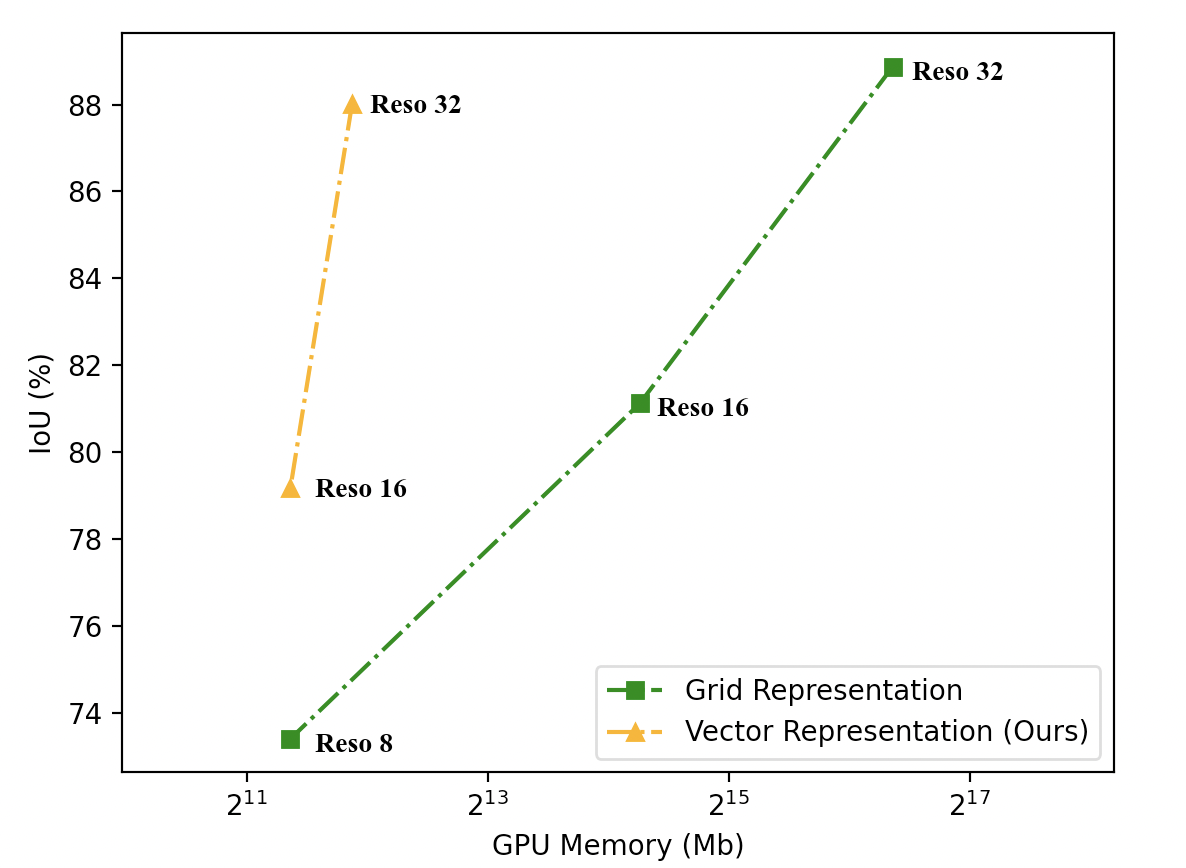}  
\tabularnewline
\vspace{-0.08in}
\caption{Comparisons of the first-stage IoU (\%) accuracy and the second-stage memory cost (Mb) with different resolutions and discrete representations. The memory cost is calculated with a batch size of 1.
\label{fig:grid_plane}}
\vspace{-0.1in}
\end{centering}
\end{figure}

\subsection*{A.3. Efficacy of Improved Discrete Representation \label{sec:ablation_order}}

In this section, we further evaluate the advantage of our improved discrete representation in terms of efficacy and efficiency.
We explore two variants of our full model by flattening coupled feature maps into vectors with row-major or column-major order. 
In Tab.~\ref{tab:uncond}, our proposed method achieves similar well performance even with different serialization orders. 
Figure~\ref{fig:order_hist} shows that our standard deviation is significantly lower than `Tri-Plane', demonstrating 
the robustness of our improved representation to generation orders. 
The proposed coupling network has facilitated AR learning by introducing more tractable order.
Additionally, the overall quality of synthesized shapes for all categories are balanced and excellent across all metrics, indicating the superiority of our design.

Furthermore, we also investigate the advantage of low computation overhead. We use `Vector' to denote our design since we apply vector quantization to latent vector, and `Grid' refers to the baseline method that applies vector quantization to volumetric grids. Figure~\ref{fig:grid_plane} compares the performance of IoU at the first stage and the corresponding memory cost in the second stage. Since we cannot afford the training of transformers with volumetric grid representations, as an alternative, we report the first-stage IoU accuracy for comparisons. From Fig.~\ref{fig:grid_plane}, two conclusion can be drawn. \textbf{(1)} The resolution $r$ of feature grids (`Grid' and `Vector') significantly affects the quality of reconstructed shapes. If $r$ is too small, it lacks the capacity to represent intricate and detailed geometries (Grid Reso-32: $88.87$ \textit{v.s} Grid Reso-16: $81.12$). However, if $r$ is large, it will inevitably increase the computational complexity in the second stage, since the number of required codes explodes as $r$ grows (Grid Reso-32: $\geq 80$G  \textit{v.s} Grid Reso-16: $19.6$G).
\textbf{(b)} Our proposed `Vector' representation not only achieves comparable reconstruction results (Vector Reso-32: $88.01$ \textit{v.s} Grid Reso-32: $88.87$), but also significantly reduces the computation overhead (Vector Reso-32: $3.8$G \textit{v.s} Grid Reso-32: $\geq 80$G).

\subsection*{A.4. Inference Speed \label{sec:speed}}
For unconditional generation, the wall time ImAM takes to sample a single shape is roughly 14 seconds, and we can also generation 32 shapes in parallel in 3 minutes.

\section*{B. Technical Details on ImAM}
\subsection*{B.1. Model Architectures \label{sec:architecture}}

\begin{figure*}[t]
\begin{centering}
\includegraphics[scale=0.4]{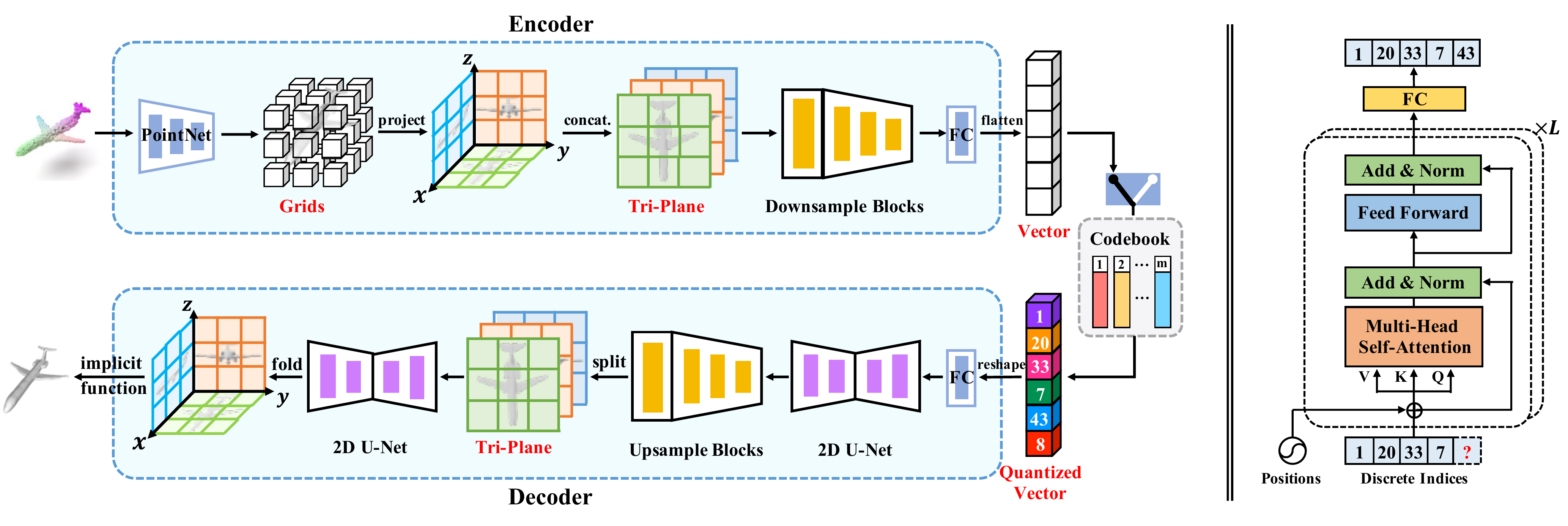}
\caption{The architectures of ImAM, consisting of an auto-encoder at the first stage (left) and a transformer at the second stage (right).
\label{fig:pipeline}}
\end{centering}
\end{figure*}

Our proposed framework ImAM consists of a two-step procedure for 3D shape generation. The first step is an auto-encoder structure, aiming to learn discrete representations for input 3D shapes. And the second step introduces a transformer structure to learn the joint distribution of discrete representations. Below we will elaborate the details of these two structures.

\noindent \textbf{Auto-encoder.} As shown in the left of Fig.~\ref{fig:pipeline}, the auto-encoder takes as input point clouds $\mathcal{P} \in \mathbb{R}^{n \times 3}$ with $n$ means the number of points, and outputs the predicted 3D mesh $\mathcal{M}$. More concretely, the encoder starts by feeding point clouds into a PointNet \cite{riegler2017octnetfusion} with local pooling, results in point features with dimensions $\mathbb{R}^{n \times 32}$. Then, we project points on three axis-aligned orthogonal planes with resolution of $256$. Features of points falling into the same spatial grid cell are aggregated via mean-operation, so that input point clouds are represented as tri-planar features instead of volumetric features. To further improve the representation, we concatenate three feature planes and couple them with three convolution layers. Next, four stacked convolution layers are adopted to not only down-sample the feature resolution three times, but also highly encode and abstract the position mapping of each spatial grid in 3D space. Thus, the output has a tractable order to be serialized as a feature vector. Before we perform the vector quantization on the flattened outputs, we follow \cite{yu2021vector} to utilize the strategy of low dimensional codebook lookup, by squeezing the feature dimension from $256$ to $4$. Consequently, an arbitrary 3D shape can be represented with a compact quantized vector, whose elements are indices of those closest entries in the codebook.

The decoder is composed of two 2D U-Net modules and one symmetric upsample block. After reshaping the quantized vector and unsqueezing its feature dimension from $4$ to $256$, we apply a 2D U-Net module to complement each spatial grid feature with global knowledge. Subsequently, the same number of 2D convolution layers as the downsmaple block are appended to upsample the feature resolution back to $256$. Symmetric convolution layers further decouple it into tri-planer features. To further improve the smoothness between the spatial grids in each plane, we use the other shared 2D U-Net module to separately process tri-plane features. The structures of both 2D U-Net are in alignment with \cite{mescheder2019occupancy}. Finally, we build a stack of fully-connected residual blocks with $5$ layer, as implicit function, to predict the occupancy probability of each query position.

\noindent \textbf{Transformer.} Benefiting from the compact discrete representation with a tractable order for each input 3D shape, we adopt a vanilla decoder-only transformer without any specific-designed module to learn the joint distributions among discrete codes. Our transformer consists of $T$ decoder layers, each of which has one multi-head self-attention layer and one feed-forward network. The decoder layer has the same structure as \cite{esser2021taming}, and is illustrated in the right of Fig.~\ref{fig:pipeline}. Specifically, we use a learnable start-of-sequence token ([SOS] token) to predict the first index of the discrete vector, and auto-regressively predict the next with the previous ones. For example, given an input containing the first $t$ indices along with one [SOS] token, we first use an embedding layer to encode them as features. Then, we feed them into several transformer layers. The position embedding is also added to provide positional information. At the end of the last transformer layer, we use two fully-connected layers to predict the \textit{logit} of each token. However, we only keep the last one which is a meaningful categorical distribution for the next ($t$+1)-th index.

\noindent \textbf{Implementation Details.} Table~\ref{tab:implentation} summarizes all parameter settings for both auto-encoder and transformer structures. We apply them as default to all experiments unless otherwise stated. 
The feature dimension $d$ in the transformer varies for different tasks. We set $d=512$ for unconditional generation; $768$ for class-guide generation and partial point completion; and $1024$ for image- or text-guide generation. We set the number of transformer layers $T=12$ for all tasks, except for text-guide generation, where we set $L=24$ and $h=16$.
Lower triangular mask matrix is used in all multi-head self-attention layers to prevent information leakage, that is, the prediction of the current index is only related to the previous known indices. 
For various conditioning inputs, we adopt the most common way to encode them. 
For example, we use a learnable embedding layer to get the feature $\in \mathbb{R}^{1\times768}$ of each category.
Gvien partial point clouds, our proposed auto-encoder encodes them into discrete representations $\in \mathbb{R}^{1024\times1}$, which are fed into another embedding layer to get features with $768$ dimensions. We adopt pre-trained CLIP models to extract features $\in \mathbb{R}^{1\times512}$ for images or texts, and further use one fully-connected layer to increase the dimension from $512$ to $1024$; All of encoded conditioning inputs are simply prepended to [SOS] token via concatenation to guide the generation.

\begin{table}[t]
\centering
\small
\setlength{\tabcolsep}{3mm}{
\caption{The detailed architecture of our framework. `k', `s' and `p' denote kernel size, stride and padding, respectively, in the convolution layer. `h8' means the number of head is 8 in multi-head self-attention layer. The feature dimension $d$ in the transformer varies for different tasks. $m$ stands for the dimension of middle layer in the feed-froward network. `K' and `L' are the sequence length of conditioning inputs and discrete representation, `1' indicates the length of [SOS] token. 
\label{tab:implentation}}
\vspace{-0.1in}
\begin{tabular}{lll}
\midrule
\multicolumn{1}{c}{Layer Name} & \multicolumn{1}{c}{Notes} & \multicolumn{1}{c}{Input Size} \tabularnewline 
\midrule
\midrule
\textbf{Auto-encoder} & & \tabularnewline 
PointNet  &  & $n\times 3$ \tabularnewline 
Coupler & &  \tabularnewline
\quad ConvLayer & k3s1p1 & $256\times256\times3\times32$ \tabularnewline 
\quad ConvLayer & k3s1p1 & $256\times256\times96$ \tabularnewline 
\quad ConvLayer & k1s1p0 & $256\times256\times32$ \tabularnewline 
Downsampler & &  \tabularnewline 
\quad ConvLayer & k2s2p0 & $256\times256\times32$ \tabularnewline 
\quad ConvLayer & k2s2p0 & $128\times128\times64$ \tabularnewline 
\quad ConvLayer & k2s2p0 & $64\times64\times128$ \tabularnewline 
\quad ConvLayer & k1s1p0 & $32\times32\times256$ \tabularnewline 
Squeezer & k1s1p0 & $32\times32\times256$ \tabularnewline 
Quantizer & & $32\times32\times4$ \tabularnewline 
Unsqueezer & k1s1p0 & $32\times32\times4$ \tabularnewline 
2D U-Net & & $32\times32\times256$  \tabularnewline 
Upsampler & &  \tabularnewline 
\quad DeconvLayer & k3s1p1 & $32\times32\times256$ \tabularnewline 
\quad DeconvLayer & k3s1p1 & $64\times64\times128$ \tabularnewline 
\quad DeconvLayer & k3s1p1 & $128\times128\times64$ \tabularnewline 
\quad ConvLayer & k1s1p0 & $256\times256\times32$ \tabularnewline 
Decoupler & &  \tabularnewline
\quad ConvLayer & k3s1p1 & $256\times256\times32$ \tabularnewline 
\quad ConvLayer & k3s1p1 & $256\times256\times96$ \tabularnewline 
\quad ConvLayer & k1s1p0 & $256\times256\times96$ \tabularnewline 
2D U-Net & & $256\times256\times3\times32$  \tabularnewline
\midrule
\midrule
\textbf{Transformer} & & \tabularnewline 
Embedding Layer &  & $\left(1+\text{L}\right) \times 1$ \tabularnewline 
Decoder Layers $\times$ 12 & &  \tabularnewline 
\quad Self-Attention & h8 & $\left(\text{K}+1+\text{L}\right) \times d$ \tabularnewline 
\quad Feed-Forward & m$4d$ & $\left(\text{K}+1+\text{L}\right) \times d$ \tabularnewline 
Head Layer & &  \tabularnewline 
\quad LinearLayer &  & $\text{L} \times d$ \tabularnewline 
\quad LinearLayer &  & $\text{L} \times d$ \tabularnewline 
\midrule
\end{tabular}}
\vspace{-0.2in}
\end{table}

\subsection*{B.2. Training and Testing Procedures \label{sec:train_test}}

\noindent \textbf{Training:} All models are trained on a single NVIDIA 3090 or A100, without any learning rate decay strategy. For the first stage, we take dense point clouds with $n=30,000$ as input, and train the auto-encoder with 13 categories on ShapeNet dataset for total 600k iterations. The learning rate is set as 1e-4, and the batch size is $16$. Once trained, it is shared for all generation tasks. 
For the second stage, we adopt the same learning rate to train the transformer. Except for the partial point completion which has the batch size of 2, we set the batch size of 8 for the other generation tasks. Models for all experiments are trained for around 600k iterations.

\noindent \textbf{Testing:} During inference, we first use the well-trained transformer to predict discrete index sequences with or without conditioning inputs. For each index, we sample it with the multinomial distribution according to the predicted probability, where only the top-\textit{k} indices with the highest confidence are kept for sampling. We progressively sample the next index, until all elements in the sequence are completed. Then, we feed the predicted index sequence into the decoder to get tri-planar features. Subsequently, we interpolate feature of each point on a grid of resolution $128^3$ from tri-planar features, and adopt the implicit function to query the corresponding occupancy probability. Finally, the iso-surface of 3D shapes are extracted with threshold of $0.2$ via Marching Cubes \cite{lorensen1987marching}.

\subsection*{B.3. Data Preparation \label{sec:data}}
We give a more detailed explanation of data preparation. We conduct all the experiments on ShapeNet dataset \cite{chang2015shapenet}. Following previous works \cite{chen2019learning,peng2020convolutional,mescheder2019occupancy}, we use 13 classes of ShapeNet dataset from 3D-R2N2 \cite{choy20163d}. The data are processed similarly to C-OccNet \cite{peng2020convolutional}. Following the same setting from IM-GAN\cite{chen2019learning} and GBIF\cite{ibing20213d}, for each category, we sorted the shapes by name and select the first 80\% as training samples and the rest for testing. 
For the task of shape completion, partial point clouds are obtained following the similar strategy in \cite{yu2021pointr}. Specifically, during training, we randomly select a viewpoint and then remove the $25\sim75\%$ furthest points from the viewpoint to obtain partial point clouds. $2048$ points are further sampled to guarantee the fixed number of partial points as inputs. For a fair comparison, we follow AutoSDF \cite{mittal2022autosdf} to devise a new completion setting, by removing all the points from the top half of shapes. For the task of image-guide generation, we use rendered images provided by 3D-R2N2\cite{choy20163d}. For text-guide generation task, we use Text2Shape dataset \cite{chen2018text2shape} with the same data splitting.

\subsection*{B.4. Implementation of Competitors \label{sec:competitors}}
We select several representative works to verify the effectiveness of our method on five tasks (\textit{i.e.}, T1: unconditional generation, T2: class-guide generation, T3: partial point completion, T4: image-guide generation, T5: text-guide generation). These works follow the priority criteria such as whether they have similar motivation, whether they conduct similar tasks, whether they release source codes, and so on. Here, we list all codebases used in our paper.
\begin{itemize}
\item IM-GAN\hspace{0.05in}\cite{chen2019learning}\hspace{0.05in}(T1):\hspace{0.05in} \url{https://github.com/czq142857/implicit-decoder}
\vspace{-0.05in}
\item GBIF \cite{ibing20213d} (T1,T2): \url{https://gitlab.vci.rwth-aachen.de:9000/mibing/localizedimplicitgan}
\vspace{-0.05in}
\item PointFlow \cite{yang2019pointflow} (T1): \url{https://github.com/stevenygd/PointFlow}
\vspace{-0.05in}
\item ShapeGF \cite{cai2020learning} (T1): \url{https://github.com/RuojinCai/ShapeGF}
\vspace{-0.05in}
\item PVD \cite{zhou20213d} (T1,T3): \url{https://github.com/alexzhou907/PVD}
\vspace{-0.05in}
\item AutoSDF \cite{mittal2022autosdf} (T2,T3,T4,T5): \url{https://github.com/yccyenchicheng/AutoSDF}
\vspace{-0.05in}
\item cGAN \cite{wu2020multimodal} (T3): \url{https://github.com/yccyenchicheng/AutoSDF}
\vspace{-0.05in}
\item ShapeFormer \cite{yan2022shapeformer} (T3): \url{https://github.com/QhelDIV/ShapeFormer}
\vspace{-0.05in}
\item Clip-Forge \cite{sanghi2022clip} (T4,T5): \url{https://github.com/AutodeskAILab/Clip-Forge}
\vspace{-0.05in}
\item ITG\hspace{0.05in}\cite{liu2022towards}\hspace{-0.15in}
(T4):\hspace{0.02in}\begin{footnotesize}\url{https://github.com/liuzhengzhe/Towards-Implicit-Text-Guided-Shape-Generation}\end{footnotesize}
\end{itemize}

Among them, we slightly modify codes of GBIF and AutoSDF for class-guide generation. Specifically, GBIF designs a generator to synthesize shape embeddings from a random noise, and a discriminator to determine whether the input embedding is real or fake. Thus, we improve it into a conditional GAN by additionally adding a class embedding as input to both the generator and discriminator. AutoSDF is also an auto-regressive model, it has the same two stages as our ImAM, including an auto-encoder and a transformer. Similar to our model, we prepend a class token to the [SOS] token, making AutoSDF be able to perform class-guide generation. Except for these two cases, we either use their provided models or re-train them with official codes if necessary for fair comparisons.

\begin{figure*} 
\begin{centering}
\includegraphics[scale=0.35]{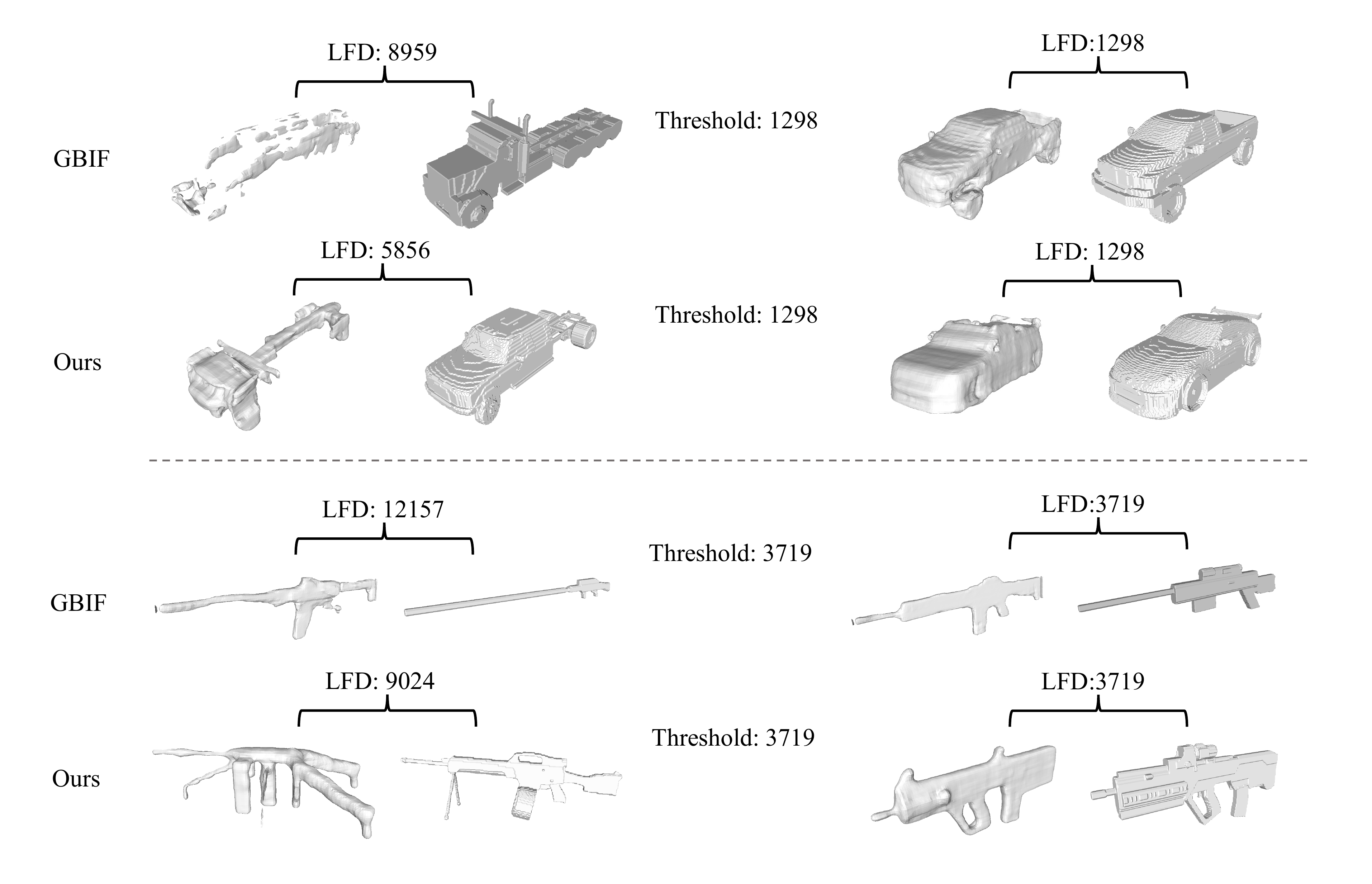}
\caption{We show the matched pairs with largest MMD before and after using threshold shifting when computing the metric of \textit{Coverage}. For each pair, the left is the generated shape and the right is the closest ground-truth shape.
\label{fig:t_cov}}
\end{centering}
\end{figure*}

\section*{C. Evaluation Metrics}

\subsection*{C.1. Implementation Details \label{sec:metric}}

As a generator, the key to evaluate our proposed method is not only to measure the \textit{fidelity}, but also to focus on the \textit{diversity} of the synthesized shapes. Therefore, we adopt eight metrics for different generation tasks, including Coverage (COV) \cite{achlioptas2018learning}, Minimum Matching Distance (MMD) \cite{achlioptas2018learning}, Edge Count Difference (ECD) \cite{ibing20213d} and 1-Nearest Neighbor Accuracy (1-NNA) \cite{yang2019pointflow}, Total Mutual Difference (TMD) \cite{wu2020multimodal}, Unidirectional Hausdorff Distance (UHD) \cite{wu2020multimodal}, Fr$\acute{\text{e}}$chet Point Cloud distance (FPD) \cite{shu20193d} and Accuracy (Acc.) \cite{sanghi2022clip}. 
In particular, we use the Light Field Descriptor (LFD) \cite{chen2003visual} as our primary similarity distance metric for COV, MMD and ECD, as suggested by \cite{chen2019learning}. Since both FPD and Acc. metrics require a classifier to calculate, we thus train a PointNet \footnote{\url{https://github.com/jtpils/TreeGAN/blob/master/evaluation/pointnet.py}} with 13 categories on ShapeNet datasets, which achieves the classification accuracy of 92\%.

For shape generation task, COV and MMD measures the diversity and fidelity of the generated shapes, respectively. Both suffer from some drawbacks \cite{yang2019pointflow}. FPD and Acc. measures the fidelity of the generated shapes from the viewpoint of feature space and probability, respectively. On the contrary, ECD and 1-NNA measure the distribution similarity of a synthesized shape set and a ground-truth shape set in terms of both diversity and quality. Therefore, ECD and 1-NNA are two more reliable and important metrics to quantify the shape generation performance.  For shape completion tasks, TMD is meant to the diversity of the generated shapes for a partial input shape, and UHD is proposed to evaluate the completion fidelity. Both metrics are specifically designed for the completion task \cite{wu2020multimodal}.

\subsection*{C.2. \textit{Coverage} with threshold \label{sec:covt}}
As discussed above, \textit{Coverage} \cite{achlioptas2018learning} measures the diversity of a generate shape set. However, it doesn't penalize outliers since a ground truth shape is still considered as covered even if the distance to the closest generated shape is large \cite{yang2019pointflow}. To rule out the false positive coverage, we count as match between a generation and ground truth shape only if LFD \cite{chen2003visual} between them is smaller than a threshold \textit{t}. In practice, \textit{t} could vary across different semantic categories based on the scale and complexity of the shape, and we empirically use MMD \cite{achlioptas2018learning} as the threshold. In this paper, we set \textit{t} as mean MMD of all competitors.

To evaluate the effectiveness of the improved COV in identifying correct matches, we visualize the matched pairs with the largest MMD before and after using threshold filtering. As shown on the left of Fig.~\ref{fig:t_cov}, when there is no threshold constraining the quality of the generated shape, outliers (\textit{e.g.}, a messy shape) could match any possible ground-truth shape, which is clearly unreasonable. On the contrary, when the threshold is applied for filtering, as illustrated on the right of Fig.~\ref{fig:t_cov}, the generated shape has certain similarity in texture or parts with the matched ground-truth shape, even if they have the maximum shape distance. It strongly demonstrates the validity and reliability of our improvement in measuring the diversity of generation.

\begin{table}[t]
\centering
\footnotesize 
\setlength{\tabcolsep}{1mm}{
\caption{Results of class-guide generation. Models are trained on 13 categories of ShapeNet. 
\label{tab:class_cond}}
\vspace{-0.1in}
\begin{tabular}{c|l|c|c|c|c|c|c}
\hline
\multirow{2}{*}{\textsc{Metrics}} &  \multicolumn{1}{c|}{\multirow{2}{*}{\textsc{Methods}}} & \multicolumn{5}{c|}{\textsc{Categories}} & \multirow{2}{*}{\textsc{AVG}}  \tabularnewline
\cline{3-7}
 & & Plane & Car & Chair & Rifle  & Table & \tabularnewline 
\hline
\hline
\multirow{3}{*}{COV $\uparrow$} 
& GBIF \cite{ibing20213d} & 68.72 & 69.64 & 75.94 &   68.98 & 81.72 & 73.00 \tabularnewline
& AutoSDF \cite{mittal2022autosdf} & 46.24 & 51.63 & 62.61 &  58.59  & 66.84 & 57.18 \tabularnewline
& \textit{Ours} & \textbf{81.58} & \textbf{71.58} & \textbf{83.98} & \textbf{75.74} & \textbf{85.48} & \textbf{79.67} \tabularnewline

\hline
\multirow{3}{*}{CovT $\uparrow$} 
& GBIF \cite{ibing20213d} & 24.10 & 38.63 & 32.69 & 35.44 & 37.80 & 33.73 \tabularnewline
& AutoSDF \cite{mittal2022autosdf} & 15.43 & 38.03 & 27.82 & 34.60 & 31.22 & 29.42 \tabularnewline
& \textit{Ours} & \textbf{56.49} & \textbf{52.70} & \textbf{45.09} & \textbf{52.74} & \textbf{49.32} & \textbf{51.27} \tabularnewline

\hline
\multirow{3}{*}{MMD $\downarrow$} 
& GBIF \cite{ibing20213d} & 4736 & 1479 & 3220 & 4246 & 2763 & 3289 \tabularnewline
& AutoSDF \cite{mittal2022autosdf} & 5201 & 1477 & 3517 & 4189   & 2992 & 3475 \tabularnewline
& \textit{Ours} & \textbf{3195} & \textbf{1285} & \textbf{2871} & \textbf{3729} & \textbf{2430} & \textbf{2702} \tabularnewline

\hline
\multirow{3}{*}{ECD $\downarrow$} 
& GBIF \cite{ibing20213d} & 1327 & 2752 & 1589 & 434 & 869 & 1394 \tabularnewline
& AutoSDF \cite{mittal2022autosdf} & 5532 & 7352 & 4136 &  2510  &  6354 & 5177 \tabularnewline
& \textit{Ours} & \textbf{571} & \textbf{1889} & \textbf{419} & \textbf{196} & \textbf{285} & \textbf{672} \tabularnewline 

\hline
\multirow{3}{*}{1-NNA $\downarrow$} 
& GBIF \cite{ibing20213d} & 91.47 & 92.43 & 75.61 & 83.12 & 70.19 & 82.56 \tabularnewline
& AutoSDF \cite{mittal2022autosdf} & 89.99 & 94.23 & 83.69 &  80.91  &  81.13 & 85.99 \tabularnewline
& \textit{Ours} & \textbf{66.81} & \textbf{83.39} & \textbf{64.83} & \textbf{57.28} & \textbf{59.55} & \textbf{66.37} \tabularnewline 

\hline
\end{tabular}}
\end{table}

\begin{table}
\centering
\small
\setlength{\tabcolsep}{1.2mm}{
\caption{Results of image-guide generation with more baselines. \label{tab:image_generation}}
\vspace{-0.1in}
\begin{tabular}{lccc} 
\midrule
\multicolumn{1}{c}{\textsc{Method}} & TMD ($\times 10^2 $) $\uparrow$ & MMD ($\times 10^3$) $\downarrow$ & FPD $\downarrow$ \tabularnewline 
\midrule
\midrule
Clip-Forge \cite{sanghi2022clip} & 2.858 & 1.926 & 8.094   \tabularnewline 
AutoSDF \cite{mittal2022autosdf} & 0.859 & 2.092 & 15.092 \tabularnewline
\hline
\textit{Ours} (ViT32)  & 3.677 & 1.617 & 2.711  \tabularnewline
\textit{Ours} (ResNet) & \textbf{4.738} & 1.662 & 3.894 \tabularnewline
\textit{Ours} (CLIP)     & 4.274 & \textbf{1.590} & \textbf{1.680}  \tabularnewline 

\hline
\end{tabular}}
\end{table}

\begin{table}
\centering
\small
\caption{Quantitative results of text-guide generation. \label{tab:text_generation}}
\vspace{-0.15in}

\setlength{\tabcolsep}{1.5mm}{
\begin{tabular}{lccc} 
\midrule
\multicolumn{1}{c}{\textsc{Method}} & TMD ($\times 10^1 $) $\uparrow$ & MMD ($\times 10^3$) $\downarrow$ & Acc $\uparrow$ \tabularnewline 
\midrule
\midrule
ITG \cite{liu2022towards}  & N/A & 2.187  &  29.13  \tabularnewline 
CLIP-Forge \cite{sanghi2022clip} & 0.400 & 2.136 &  53.68  \tabularnewline 
\hline
\textit{Ours} (BERT) & \textbf{0.677} & 1.931 & \textbf{60.68} \tabularnewline
\textit{Ours} (CLIP-seq.)  & 0.524 & \textbf{1.778} & 58.17 \tabularnewline
\textit{Ours} (CLIP)    & 0.565 & 1.846 & 59.93  \tabularnewline 
\hline
\end{tabular}}
\end{table}

\section*{D. More Experimental Analysis}
\subsection*{D.1. 1-NNA metric on Class-guide Generation \label{sec:class_gen}}
Both 1-NNA and ECD are two important metrics to measure the overall quality of the synthesized shapes in terms of fidelity and diversity. Thus, we extend Tab.~\textcolor{red}{2} of the manuscript by additionally adding comparison results of 1-NNA accuracy, as shown in Tab.~\ref{tab:class_cond}. As expected, ImAM outperforms both competitors across 5 categories by a significant margin, achieving state-of-the-art results on all metrics. In particular, it also shows significant advantages on the metric of 1-NNA, strongly demonstrating the versatility of ImAM on class-guide generation.

\subsection*{D.2. More Baselines on Image-guide Generation \label{sec:image_gen}}
To further evaluate the effectiveness of ImAM on image-guide generation, we compare it with AutoSDF \cite{mittal2022autosdf}, which is also an auto-regressive model for 3D shape generation. We follow its official codes to extract per-location conditionals from the input image and then guide the non-sequential auto-regressive modeling to generate shapes. 
On the contrary, our method only utilize a single image feature extracted from ResNet or CLIP, guiding the vanilla transformer, to auto-regressively synthesize shapes.
As shown in Tab.~\ref{tab:image_generation}, ImAM beats AutoSDF on three metrics by a large margin, which clearly suggests the effectiveness of our proposed method. In addition, AutoSDF requires a specific form of conditioning inputs, but our approach can accept input in either 1- or 2-D form, giving it more flexibility. To verify it, we conduct a baseline by using ViT32 to extract patch embeddings from the input image as condition. The conditional generation can be achieved by simply prepending patch embeddings to [SOS] token. Results in Tab.~\ref{tab:image_generation} indicate that it is competitive with models using ResNet or CLIP as feature extractor, 
further suggesting the powerful versatility of our ImAM in either generative ways or conditional forms.

\subsection*{D.3. More Baselines on Text-guide Generation \label{sec:text_gen}}
Different images, texts have a natural form of sequence. Each word is closely related to its context. Thereby, we further discuss the ability of ImAM to text-guide generation conditioned on sequence embeddings. Concretely, we adopt BERT and CLIP \footnote{In this case, we output features of all tokens instead of the last [END] token as the text embedding.} model to encode texts into a fixed length of sequence embeddings. From Tab.~\ref{tab:text_generation}, we find that using sequence embeddings indeed boost the performance of ImAM for text-guide generation task. Thanks to our proposed compact discrete representation with more tractable orders, our ImAM can be easily adapted to different conditional forms for different tasks, thus improving the quality of the generated shapes.

\section*{E. More Qualitative Results}

\subsection*{E.1. More Visualizations of Generated Shapes}
\begin{figure*}[t]
\begin{centering}
\includegraphics[width=17cm, height=21cm]{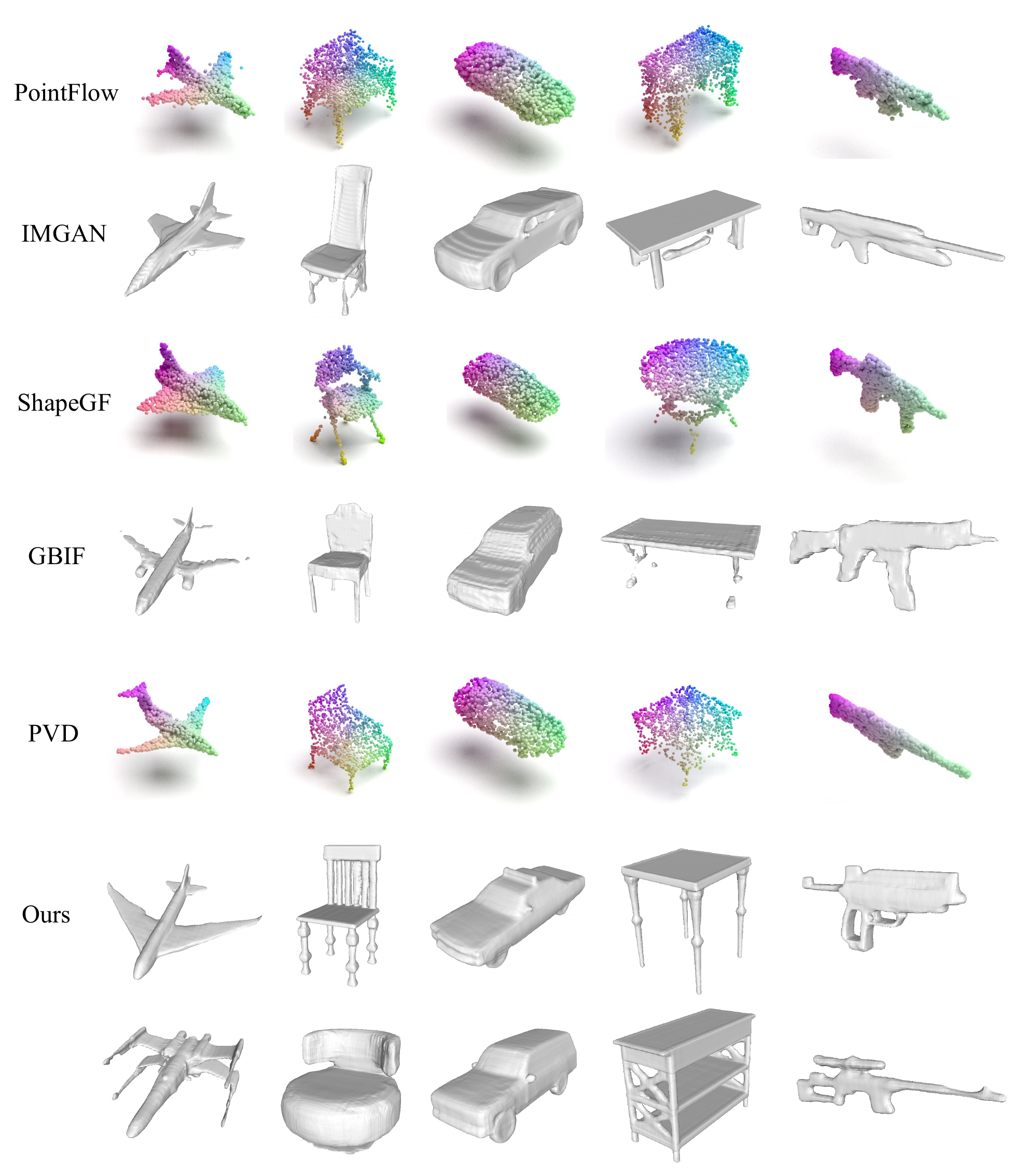}
\caption{More qualitative comparisons of unconditional generation on 5 categories. \label{fig:uncond1}}
\end{centering}
\end{figure*}

\begin{figure*}[t]
\begin{centering}
\includegraphics[scale=0.71]{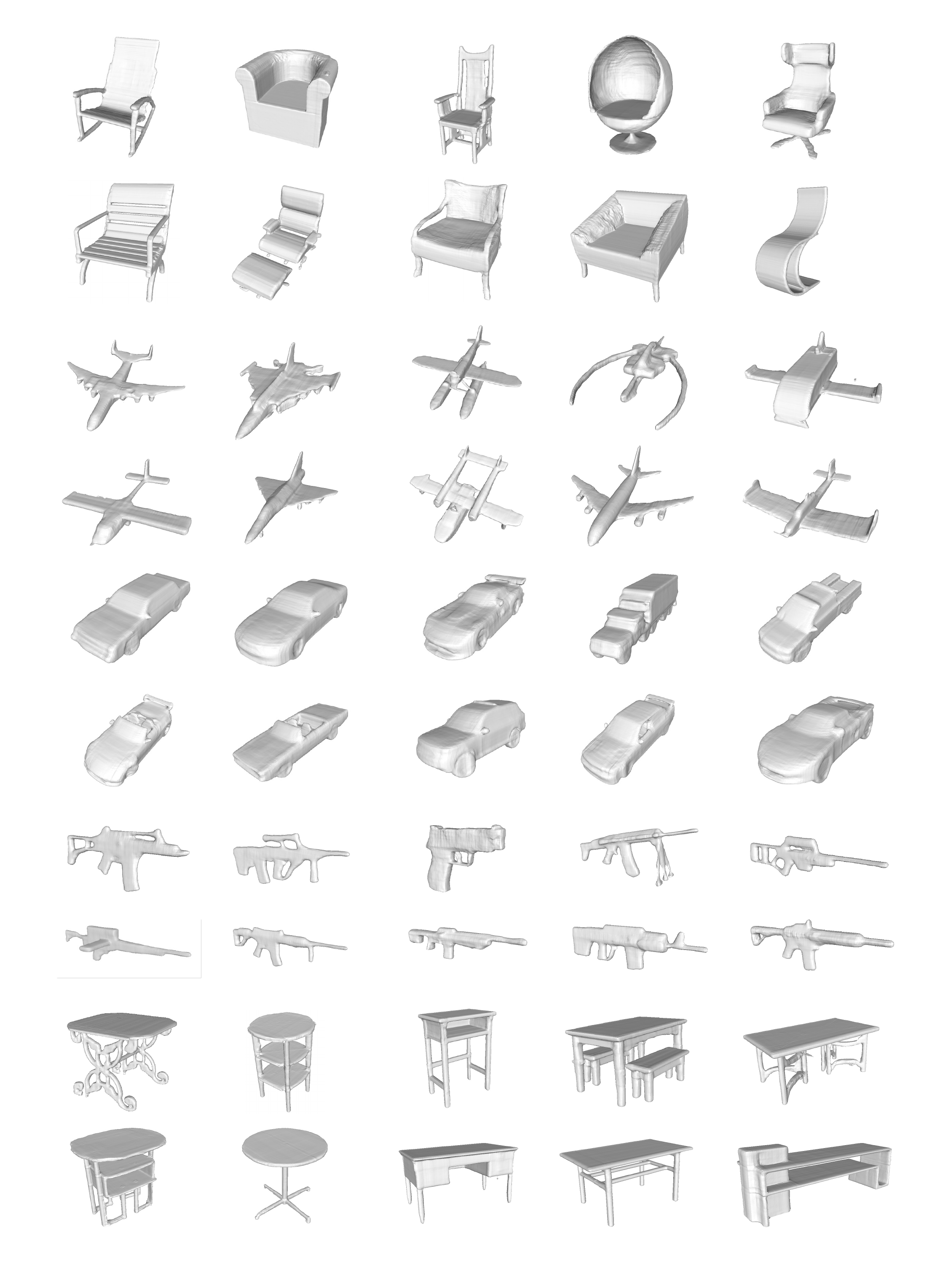}
\vspace{-0.15in}
\caption{More visualization of unconditional generation on 5 categories. \label{fig:uncond2}}
\end{centering}
\end{figure*}

We show qualitative comparisons of unconditional generation in Fig.~\ref{fig:uncond1}. Our ImAM can generate more faithful and diverse shapes of multiple categories. Comprehensive visualizations of generated shapes are illustrated in Fig.~\ref{fig:uncond2}. ImAM can generate shapes with fine geometric structures (\textit{e.g.}, airplanes and cars), as well as some shapes with composite or hollowed-out designs (\textit{e.g.}, tables and chairs). One major key is that it enjoys the advantages of more compact discrete representations while endowing tractable orders to learn shape priors of complicated geometries.

\begin{figure*}[t]
\begin{centering}
\includegraphics[scale=0.25]{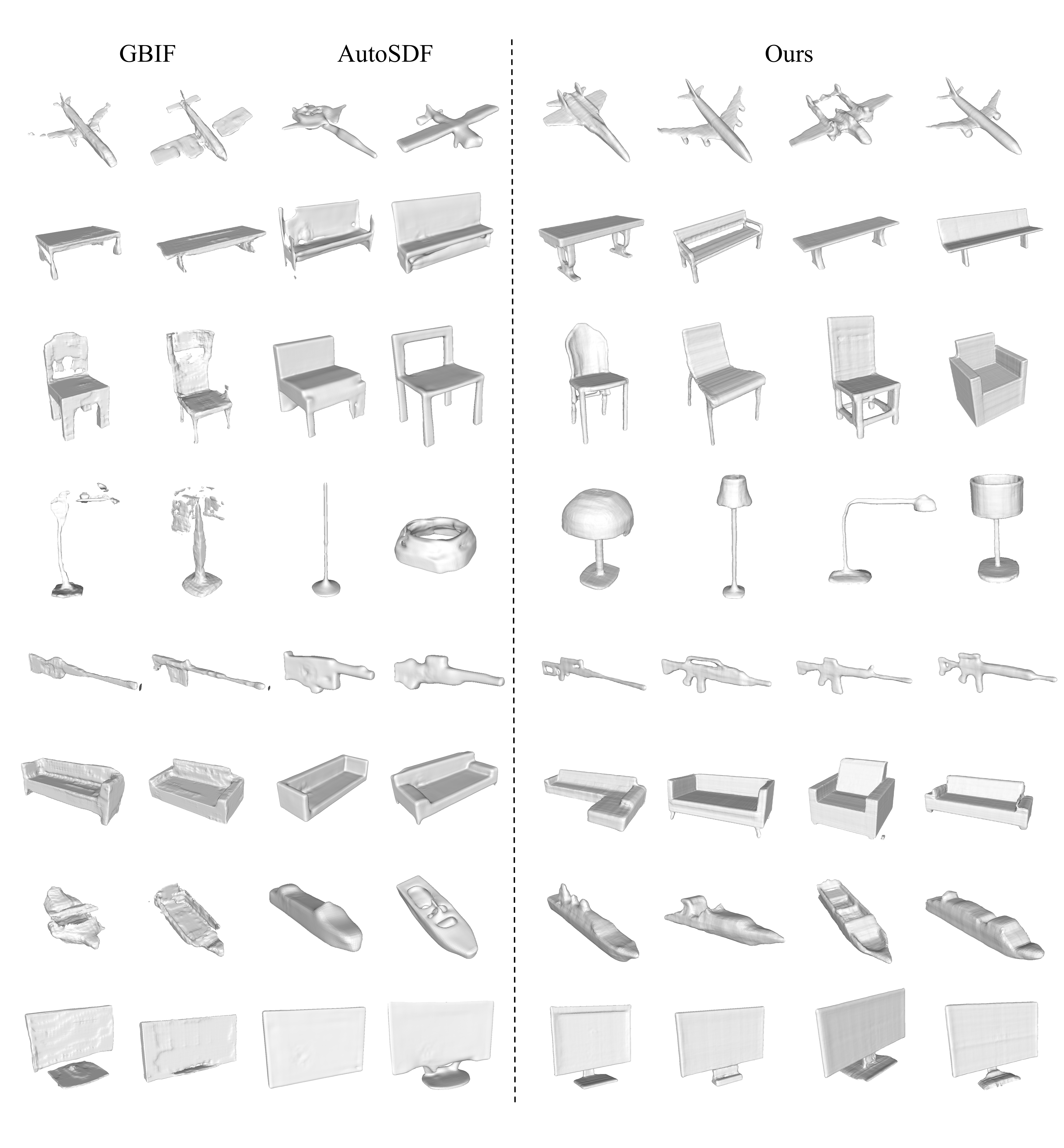}
\caption{More qualitative comparisons of class-guide generation. \label{fig:class1}}
\end{centering}
\end{figure*}

\begin{figure*}[t]
\begin{centering}
\includegraphics[scale=0.62]{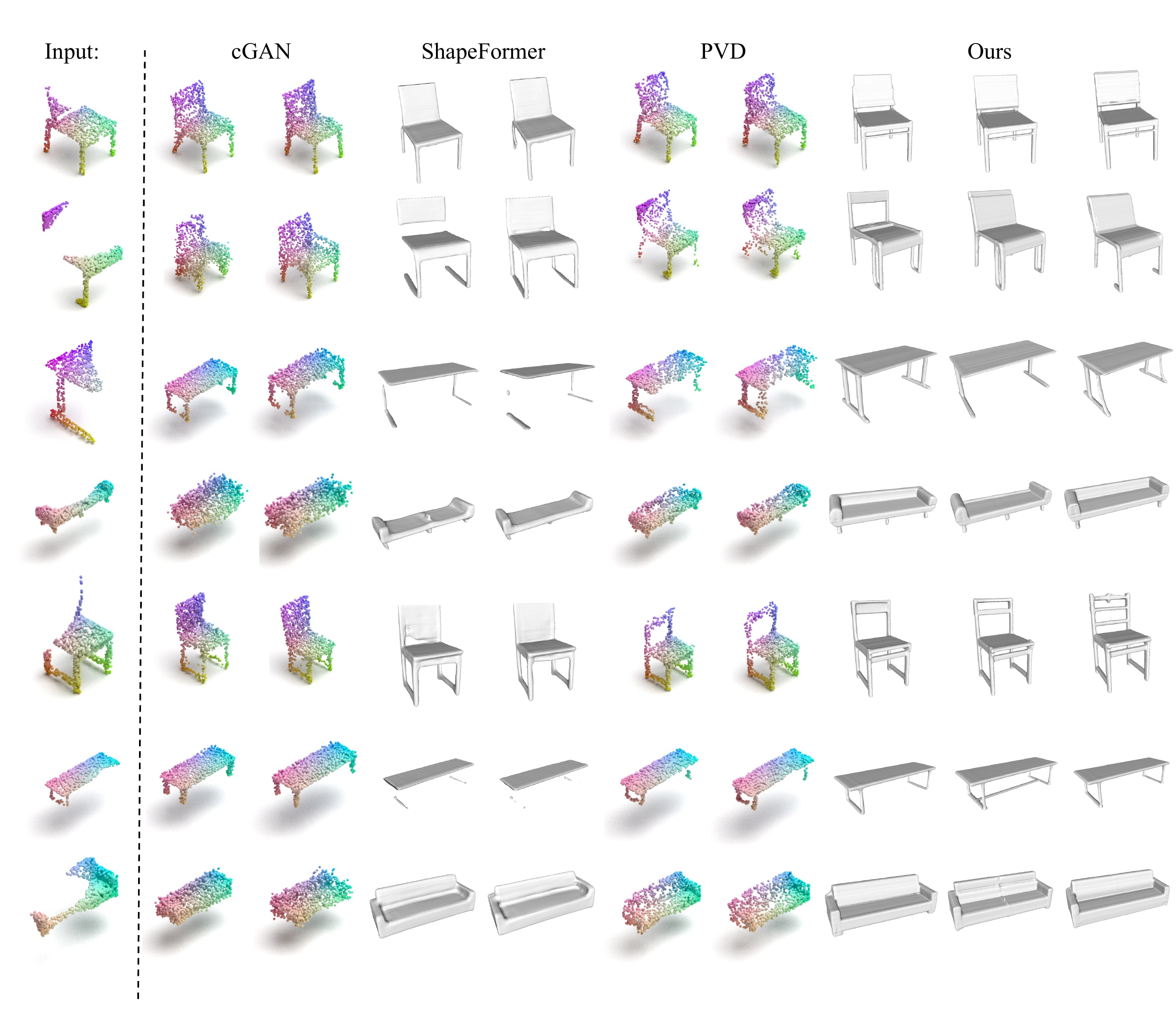}
\vspace{-0.1in}
\caption{More qualitative comparisons of multi-modal partial point completion on 3 categories. \label{fig:partial1}}
\end{centering}
\end{figure*}

\begin{figure*}[t]
\begin{centering}
\includegraphics[scale=0.80]{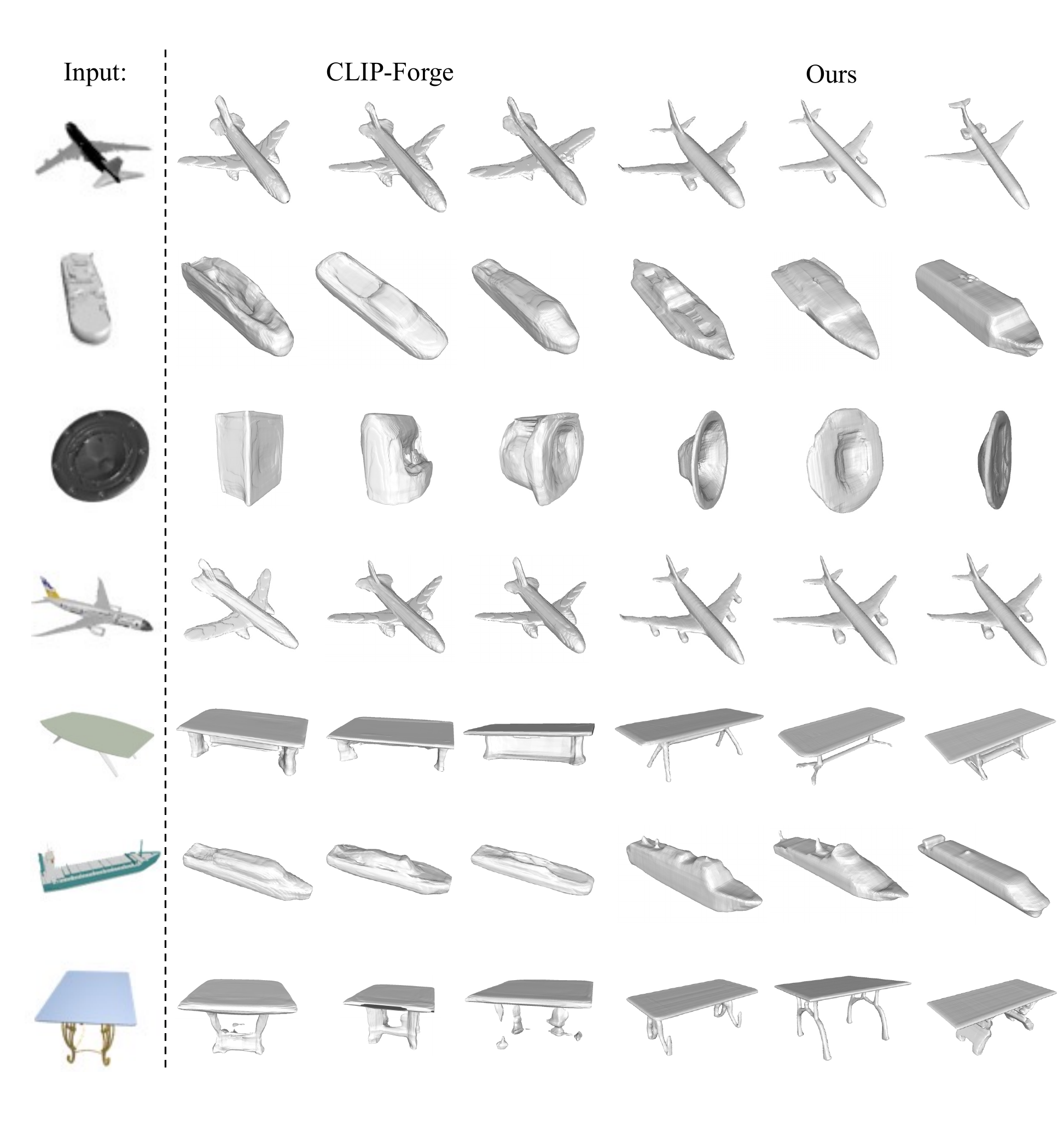}
\vspace{-0.1in}
\caption{More qualitative comparisons of image-guide generation. \label{fig:image1}}
\end{centering}
\end{figure*}

\begin{sidewaysfigure}[t]
\vspace{9cm}
\begin{centering}
\includegraphics[scale=0.8]{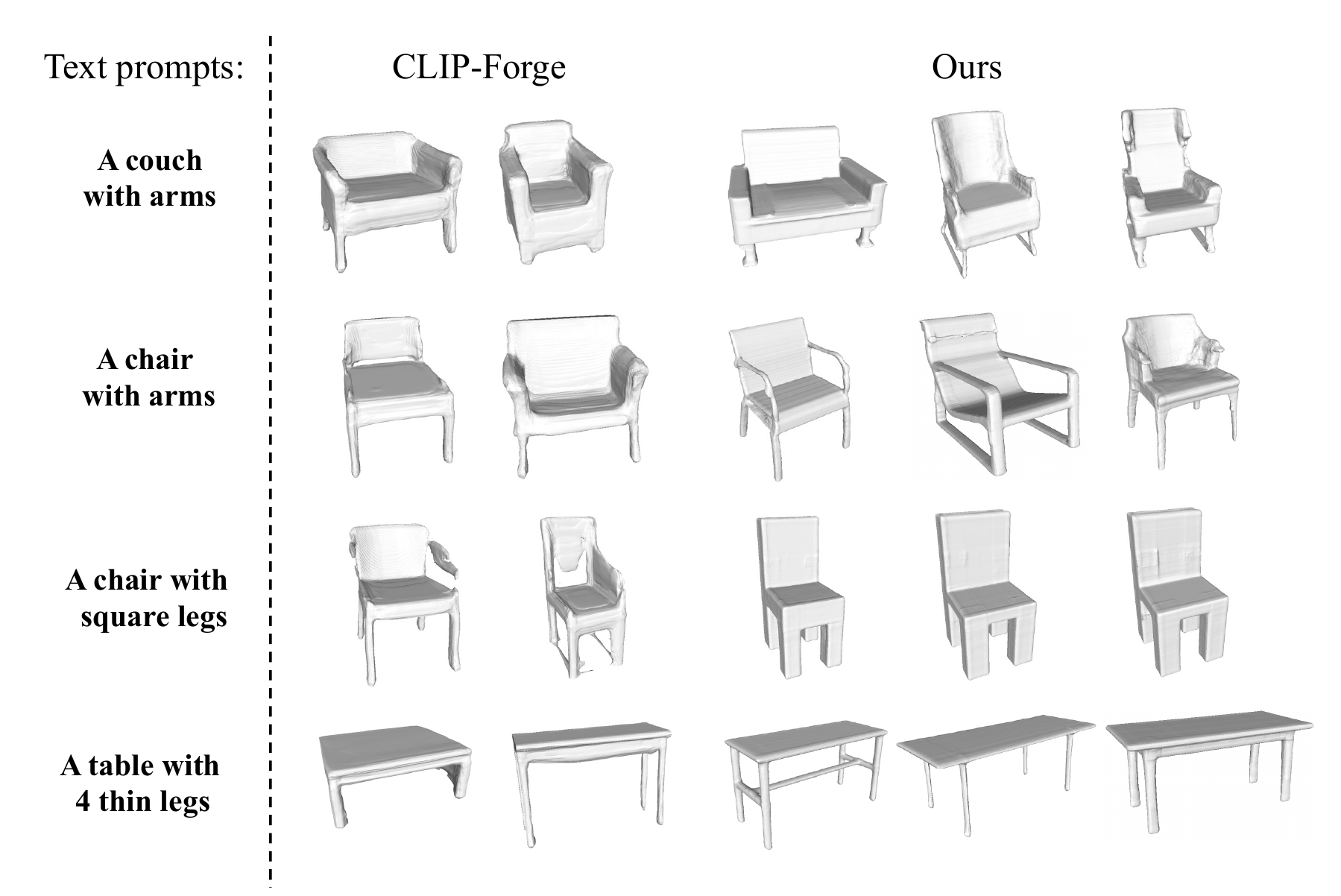}
\vspace{-0.1in}
\caption{More qualitative comparisons of text-guide generation. Models are trained on Text2Shape dataset which only includes two categories, \textit{i.e.}, chair and table. \label{fig:text1}}
\end{centering}
\end{sidewaysfigure}

Moreover, we show more synthesized samples of class-guide generation in Fig.~\ref{fig:class1}, partial point completion in Fig.~\ref{fig:partial1}, image-guide generation in Fig.~\ref{fig:image1}, and text-guide generation in Fig.~\ref{fig:text1}. 
ImAM can generates high-quality shapes not only  being faithful to the given conditions, but also showing large imagination on diversity, which significantly indicates a more unified ability to freely turn unconditional generation into conditional generation.

\subsection*{E.2. Image-guide Generation in Real-world}
\begin{figure*}[t]
\begin{centering}
\includegraphics[scale=0.8]{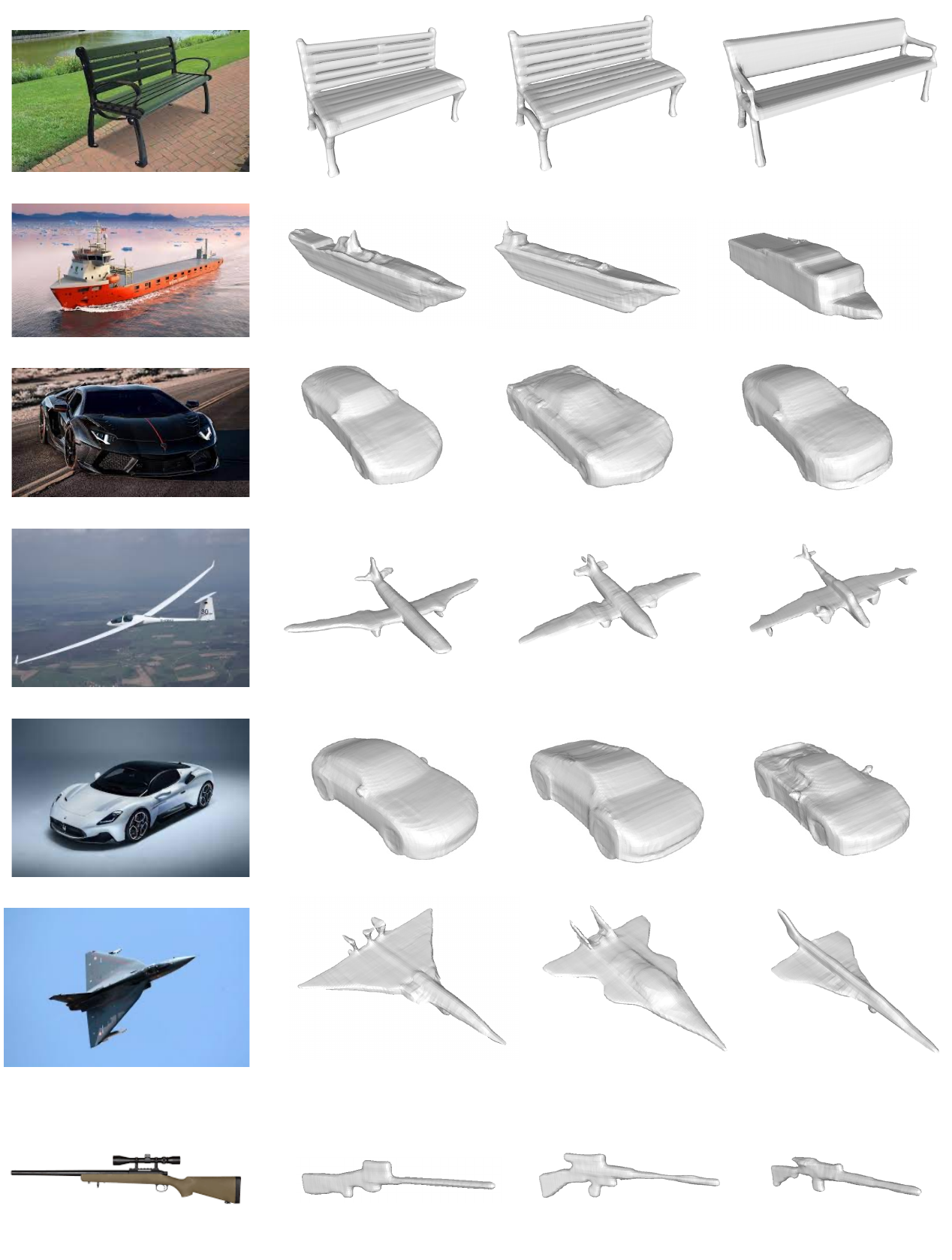}
\vspace{-0.1in}
\caption{More qualitative results of image-guide generation from real-world data. \label{fig:real-world}}
\end{centering}
\end{figure*}

To further investigate the application of ImAM on real-world image-guide generation, we show more synthesized shapes by given images captured in the real world. We use the same model as described in Sec.~\textcolor{red}{4.4} of the manuscript. As illustrated in Fig.~\ref{fig:real-world}, results on five categories suggest that our model can sensitively capture major attributes of objects in the image. 
Take the first two rows as examples, ImAM can generate plausible shapes while being faithful to objects in images. In addition, two airplane images in Fig.~\ref{fig:real-world} show that
our synthesized samples enjoy the advantage of diversity by partially sticking to the major attributes, such the types of wings and tail.

\begin{figure*}[t]
\begin{centering}
\includegraphics[scale=0.65]{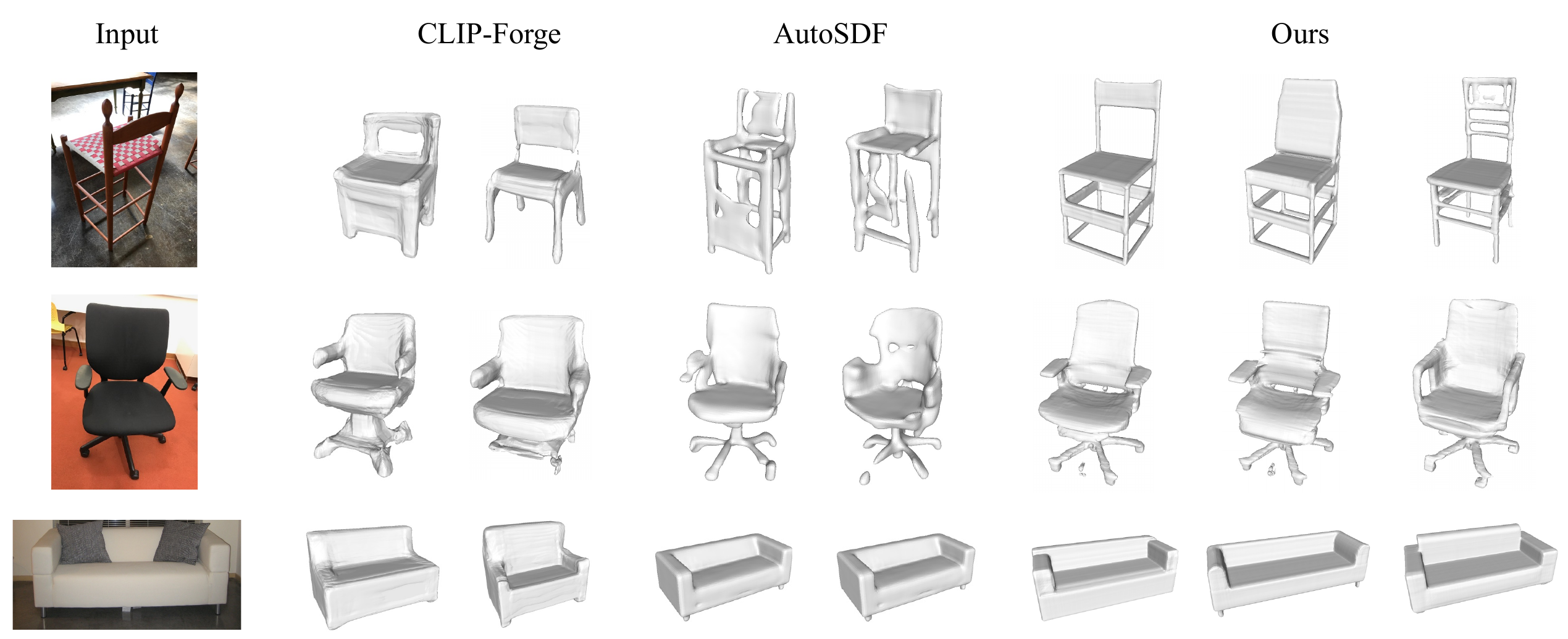}
\vspace{-0.1in}
\caption{More qualitative results of image-guide generation from real-world dataset Pix3D.  \label{fig:pix3d}}
\end{centering}
\end{figure*}

Figure~\ref{fig:pix3d} also show results of our model on Pix3D dataset  (trained on ShapeNet, without any finetuning).
Compared with other competitors, ImAM is 
capable of generating high-quality and realistic shapes that highly match the shape of objects in real-world images. It significantly highlights the strong generalization ability of our method.

\subsection*{E.3. Zero-shot Text-to-shape Generation}
\begin{figure*}[t]
\begin{centering}
\includegraphics[scale=1.1]{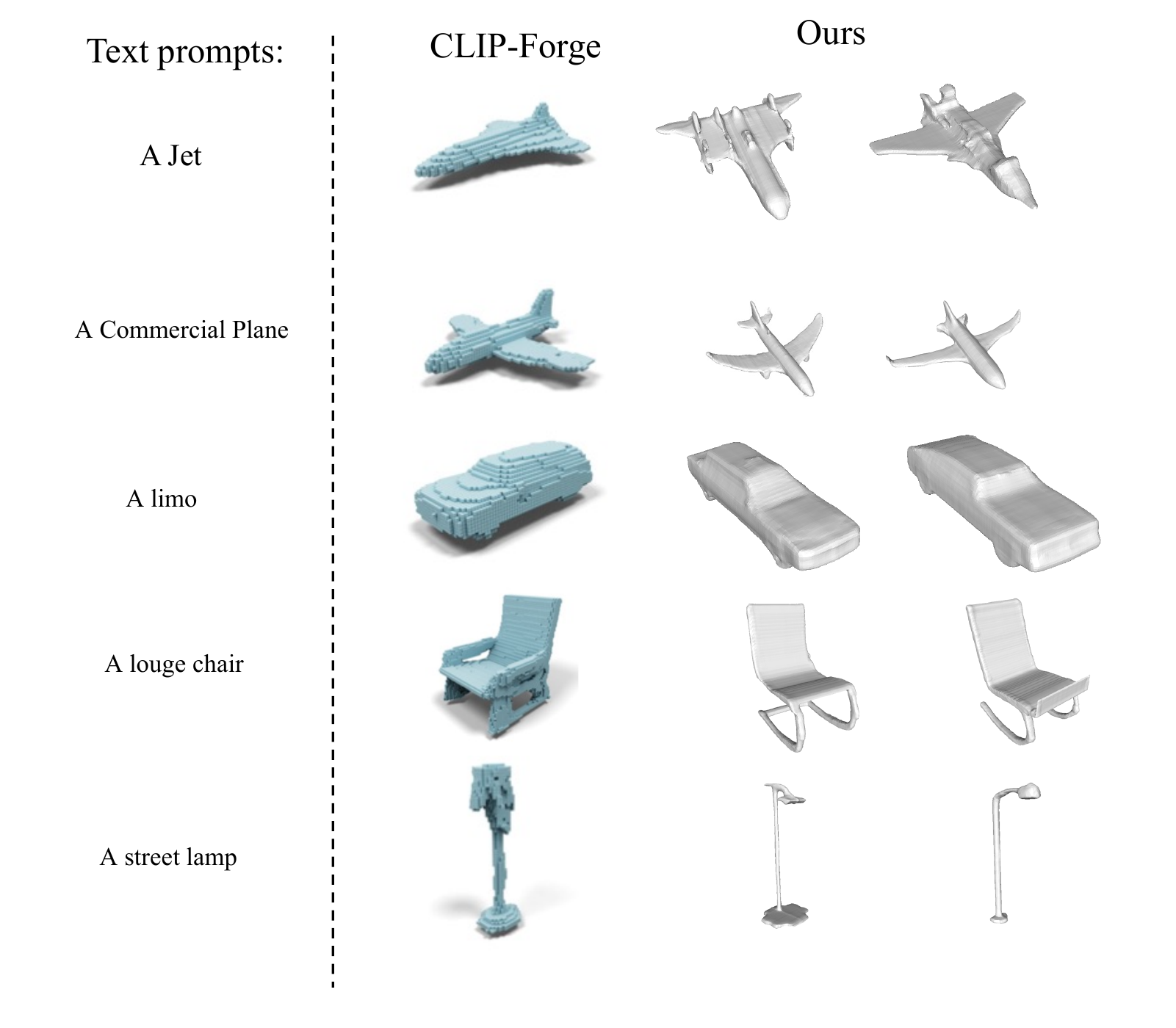}
\vspace{-0.1in}
\caption{More qualitative results of zero-shot text-to-shape generation. Results of CLIP-Froge are reported in their paper.
\label{fig:zero-shot}}
\end{centering}
\end{figure*}

Figure~\ref{fig:zero-shot} shows more qualitative results of zero-shot text-to-shape generation. Specifically, our model is trained on multiple pairs of image and shapes, and directly take  text conditions as input during inference, as the same setting in \cite{sanghi2022clip}. The high-quality of synthesized shapes clearly demonstrate the 
powerful versatility of our proposed ImAM in 3D shape generation, showing great potential to the real-world applications.

\section*{F. Broader Impact and Limitation}
\noindent \textbf{Broader Impact.} Synthesizing high quality 3D content has strong application in the fields of AR/VR, graphics, robotics and metaverse. In the past, creating high quality 3D content requires great effort from professional designers and also a great amount of time. Our work ImAM can serve as a tool for automatically generating high quality 3D shapes, providing convenience for 3D designers to create more sophisticated 3D content. However, at the same time, special care must be taken not to infringe the copyright of other 3D content creators during the process of data collection for our model training.

\noindent \textbf{Limitation.} Despite we have made huge effort in adopting an efficient 3D representation in our model, ImAM still inherit the limitation of auto-regressive model. The inference time is relatively consuming, since it requires multiple forward operations to generate one sample. Besides, the way of auto-regression may suffer from the problem of error accumulation. In particular, if conditioning inputs contain some noise, it is very fragile to synthesize incorrect shapes, or even collapsed ones. In future work, we will explore more efficient auto-regressive architectures and representations to overcome these limitations.

\newpage

\end{document}